\documentclass[10pt,twocolumn]{article}

\usepackage[T1]{fontenc}
\usepackage{mathpazo}      
\usepackage[scaled=0.9]{helvet}  
\usepackage{courier}       
\usepackage{microtype}
\usepackage{graphicx}
\usepackage{amsmath}
\usepackage{amssymb}
\usepackage{booktabs}
\usepackage{tabularx}
\usepackage{xcolor}
\usepackage{listings}
\PassOptionsToPackage{hyphens,spaces,obeyspaces}{url}
\usepackage[round,authoryear]{natbib}
\usepackage{hyperref}
\usepackage{cleveref}

\makeatletter
\g@addto@macro{\UrlBreaks}{\UrlOrds}
\makeatother
\usepackage{fontawesome5}
\newcommand{\emoji}[1]{\,\textcolor{orange}{\faFire}\,}
\usepackage{tikz}
\usetikzlibrary{shapes,arrows,positioning,shadows,calc,backgrounds,decorations.pathreplacing}
\usepackage{subcaption}
\usepackage{enumitem}
\usepackage{titlesec}
\usepackage{fancyhdr}

\newcommand{\fullmark}{\checkmark}
\newcommand{\halfmark}{$\circ$}
\newcommand{\emptymark}{--}

\titlespacing*{\section}{0pt}{0.8\baselineskip}{0.5\baselineskip}
\titlespacing*{\subsection}{0pt}{0.6\baselineskip}{0.4\baselineskip}
\titlespacing*{\subsubsection}{0pt}{0.5\baselineskip}{0.3\baselineskip}

\usepackage[
  letterpaper,
  top=0.75in,
  bottom=1in,
  left=0.75in,
  right=0.75in,
  columnsep=0.25in
]{geometry}

\definecolor{accentcolor}{RGB}{255,87,34} 

\fancypagestyle{plain}{%
  \fancyhf{}%
  \fancyhead{}%
  \fancyfoot{}%
}

\definecolor{codegreen}{rgb}{0,0.5,0}
\definecolor{codegray}{rgb}{0.4,0.4,0.4}
\definecolor{codepurple}{rgb}{0.58,0,0.82}
\definecolor{backcolour}{rgb}{0.98,0.98,0.98}

\newcommand{\boxednumber}[1]{\makebox[0.85em][r]{\fontsize{4.5}{5}\selectfont\ttfamily\color{codegray}#1}}

\lstdefinestyle{pythonstyle}{
    backgroundcolor=\color{backcolour},
    commentstyle=\color{codegreen},
    keywordstyle=\color{blue},
    numberstyle=\boxednumber,
    stringstyle=\color{codepurple},
    basicstyle=\ttfamily\scriptsize,
    breakatwhitespace=false,
    breaklines=true,
    captionpos=b,
    keepspaces=true,
    numbers=left,
    numbersep=4pt,
    showspaces=false,
    showstringspaces=false,
    showtabs=false,
    tabsize=2,
    language=Python,
    frame=single,
    framerule=0.5pt,
    rulecolor=\color{black!20},
    xleftmargin=3pt,
    xrightmargin=3pt,
    aboveskip=5pt,
    belowskip=5pt
}

\definecolor{pytorchbg}{rgb}{0.92,0.92,0.95}
\lstdefinestyle{pytorchstyle}{
    backgroundcolor=\color{pytorchbg},
    commentstyle=\color{codegreen},
    keywordstyle=\color{blue},
    numberstyle=\boxednumber,
    stringstyle=\color{codepurple},
    basicstyle=\ttfamily\scriptsize,
    breakatwhitespace=false,
    breaklines=true,
    captionpos=b,
    keepspaces=true,
    numbers=left,
    numbersep=4pt,
    showspaces=false,
    showstringspaces=false,
    showtabs=false,
    tabsize=1,
    basewidth=0.48em,
    language=Python,
    frame=single,
    rulecolor=\color{black!30},
    xleftmargin=3pt,
    xrightmargin=5pt,
    aboveskip=6pt,
    belowskip=6pt
}

\hypersetup{
    colorlinks=true,
    linkcolor=blue,
    citecolor=blue,
    urlcolor=blue,
    pdftitle={TinyTorch: Building Machine Learning Systems from First Principles},
    pdfauthor={Vijay Janapa Reddi}
}

\usepackage{titling}

\pretitle{\begin{center}\vspace*{-1em}}
\posttitle{\vspace{0.8em}\end{center}}

\title{
  \Huge\bfseries Tiny\emoji{fire}Torch\\[0.4em]
  \Large\normalfont\itshape Building Machine Learning Systems from First Principles
}

\author{
  \fontsize{12}{15}\selectfont
  Vijay Janapa Reddi\\[0.2em]
  \fontsize{11}{14}\selectfont
  Harvard University\\[1.2em]
  \fontsize{10}{12}\selectfont
  \textcolor{gray!60}{\href{https://tinytorch.ai}{tinytorch.ai}}
}

\date{}

\begin{document}

\thispagestyle{plain}
\maketitle

\begin{abstract}
  Machine learning education faces a fundamental gap: students learn algorithms without understanding the systems that execute them. They study gradient descent without measuring memory, attention mechanisms without analyzing $O(N^2)$ scaling, optimizer theory without knowing why Adam requires $3\times$ the optimizer-related memory of SGD (gradients plus two state buffers). This \emph{algorithm-systems divide} produces practitioners who can train models but cannot debug memory failures, optimize inference latency, or reason about deployment trade-offs—the very skills industry demands as ``ML systems engineering.'' We present TinyTorch, a 20-module curriculum that closes this gap through \emph{implementation-based systems pedagogy}: students construct PyTorch's core components (tensors, autograd, optimizers, CNNs, transformers) in pure Python, building a complete framework where every operation they invoke is code they wrote. The design employs three patterns: \emph{progressive disclosure} of complexity, \emph{systems-first integration} of profiling from the first module, and \emph{build-to-validate milestones} recreating 67 years of ML breakthroughs—from Perceptron (1958) through Transformers (2017) to MLPerf-style benchmarking. Requiring only 4GB RAM and no GPU, TinyTorch demonstrates that deep ML systems understanding is achievable without specialized hardware. The curriculum is available open-source at \texttt{mlsysbook.ai/tinytorch}.
  \end{abstract}

\section{Introduction}
\label{sec:intro}

In ``The Bitter Lesson,'' Rich Sutton observes that the history of artificial intelligence teaches a counterintuitive truth: general methods that leverage computation ultimately defeat cleverly-designed, domain-specific approaches~\citep{sutton2019bitter}. Deep learning surpassed handcrafted features in computer vision. Large language models outperformed linguistic rule systems. AlphaZero mastered games through self-play rather than encoded heuristics. A fundamental driver behind each breakthrough is computational efficiency: the ability to effectively scale learning systems to leverage available hardware. Yet while we have learned this lesson algorithmically, building ever-larger models that demonstrate its truth, we have not embedded it pedagogically. Our educational systems continue to separate the teaching of machine learning algorithms from the systems knowledge required to achieve computational scale.

This pedagogical gap creates a systems efficiency crisis. Modern ML models often fail not from algorithmic limitations but from hitting computational walls. Transformer attention mechanisms scale as $O(N^2)$ with sequence length~\citep{vaswani2017attention}, causing memory exhaustion before accuracy plateaus. Distributed training requires understanding gradient synchronization overhead that can eliminate parallel speedup. Production deployments crash from subtle memory leaks in tensor caching and reference cycles. The paradox is stark: we know empirically that scale wins, that computational efficiency enables the breakthroughs Sutton documents, yet we do not teach students how to achieve this scale. They can train models but cannot explain why gradient accumulation reduces memory usage or when activation checkpointing becomes necessary.

This crisis directly impacts the ML workforce. Industry surveys indicate significant demand-supply imbalances for ML systems engineers~\citep{roberthalf2024talent,keller2025ai}, with surveys suggesting that a substantial portion of executives cite talent shortage as their primary barrier to AI adoption~\citep{keller2025ai}. These are not the scientists who invent new architectures but the engineers who make existing architectures computationally viable, who understand when mixed precision training preserves accuracy, how gradient checkpointing trades compute for memory, and why distributed training introduces synchronization bottlenecks. The emerging recognition of AI engineering as a discipline distinct from both ML research and traditional software engineering~\citep{sei2020aieng} reflects this gap: bridging what research makes possible and what production deployment requires demands integrative competencies that neither field alone provides. These practitioners are often the bottleneck to realizing the promise of computational scaling that enables general methods to triumph.

The knowledge these engineers need is about systems mental models. Understanding \emph{why} Adam requires $2\times$ optimizer state memory requires visualizing optimizer state buffers. Predicting \emph{when} batch sizes must shrink to fit GPU memory requires internalizing memory hierarchy latencies. Navigating accuracy-latency-memory tradeoffs in production systems requires understanding collective communication patterns and their overhead~\citep{meadows2008thinking}. This tacit systems knowledge (how frameworks manage memory, schedule operations, and optimize execution) cannot be developed through high-level API usage alone. It emerges from building these systems, from implementing tensor operations that reveal memory access patterns, from constructing computational graphs exposing optimization opportunities.

Current ML education often fails to develop these mental models through a strict separation: algorithms courses teach backpropagation mathematics while systems courses teach distributed computing, with limited bridges between them. Students learn gradient descent's convergence properties but not its memory footprint. They implement neural networks using framework APIs but never see how those APIs translate to memory allocations and kernel launches. They can mathematically derive the chain rule but cannot explain how \texttt{autograd} implements it efficiently through dynamic graph construction and topological traversal. This separation leaves students unprepared for the systems engineering roles that the industry desperately needs.

We present TinyTorch, a 20-module curriculum that teaches single node machine learning systems engineering by building a PyTorch-compatible framework from pure Python primitives. Designed as a hands-on companion to the \emph{Machine Learning Systems} textbook~\citep{mlsysbook2025}, TinyTorch makes systems efficiency tangible: students implement tensor operations while measuring memory consumption, build autograd while profiling computational graphs, create optimizers while tracking state overhead. Each module reinforces insights inspired by the systems imperative the Bitter Lesson reveals: computational efficiency, not algorithmic cleverness alone, drives ML progress. Students don't just learn \emph{that} Conv2d achieves 109$\times$ parameter efficiency compared to dense layers, they \emph{implement} sliding window convolution and \emph{measure} the difference directly through profiling code they wrote.

This approach mirrors computer engineering education, where processor design progresses from transistors to logic gates to arithmetic units to complete CPUs, with each layer a working system that enables the next. TinyTorch applies the same principle: tensors enable operations, operations enable layers, layers enable networks, networks enable complete ML systems. \Cref{fig:code-comparison} illustrates this progression: from PyTorch's black-box APIs, through building internals like optimizers, to training transformers where every import is student-implemented code.

\begin{figure*}[t]
\centering
\hfill
\begin{minipage}[b]{0.32\textwidth}
\begin{subfigure}[b]{\textwidth}
\centering
\begin{lstlisting}[basicstyle=\fontsize{6}{7}\selectfont\ttfamily,frame=single,style=pythonstyle]
import torch.nn as nn
import torch.optim as optim

# How much memory?
model = nn.Linear(784, 10)

# Why does Adam need more
# memory than SGD?
optimizer = optim.Adam(
              model.parameters())
loss_fn = nn.CrossEntropyLoss()

for epoch in range(10):
  for x, y in dataloader:
    pred = model(x)
    loss = loss_fn(pred, y)
    loss.backward()  # Magic?
    optimizer.step() # How?
\end{lstlisting}
\vspace{0.15em}
\subcaption{PyTorch: Black box usage}
\label{lst:pytorch-usage}
\end{subfigure}
\end{minipage}
\hfill
\begin{minipage}[b]{0.31\textwidth}
\begin{subfigure}[b]{\textwidth}
\centering
\begin{lstlisting}[basicstyle=\fontsize{6}{7}\selectfont\ttfamily,frame=single,style=pythonstyle]
class Adam:
  def __init__(self, params,
               lr=0.001):
    self.params = params
    self.lr = lr
    # 2$\times$ optimizer state:
    # momentum + variance
    self.m = [zeros_like(p)
              for p in params]
    self.v = [zeros_like(p)
              for p in params]

  def step(self):
    for i, p in enumerate(
            self.params):
      self.m[i] = 0.9*self.m[i]
                + 0.1*p.grad
      self.v[i] = 0.999*self.v[i]
                + 0.001*p.grad**2
      p.data -= self.lr *
        self.m[i] /
        (self.v[i].sqrt()+1e-8)
\end{lstlisting}
\vspace{0.15em}
\subcaption{TinyTorch: Build internals}
\label{lst:tinytorch-build}
\end{subfigure}
\end{minipage}
\hfill
\begin{minipage}[b]{0.3\textwidth}
\begin{subfigure}[b]{\textwidth}
\centering
\begin{lstlisting}[basicstyle=\fontsize{6}{7}\selectfont\ttfamily,frame=single,style=pythonstyle]
# After Module 13: train
# transformers with YOUR code
from tinytorch import (
  GPT, Embedding)
from tinytorch import Adam
from tinytorch import DataLoader

model = GPT(
 vocab=1000, d_model=64,
 n_heads=4, n_layers=2)
 opt = Adam(model.parameters())

for batch in DataLoader(data):
  loss = model(batch.x,
               batch.y)
  loss.backward() # Yours!
  opt.step()      # Yours!
  # You understand WHY it works
  # because you built it all!
\end{lstlisting}
\vspace{0.15em}
\subcaption{TinyTorch: The culmination}
\label{lst:tinytorch-culmination}
\end{subfigure}
\end{minipage}
\hfill
\caption{\textbf{From User to Engineer.} (a) PyTorch's high-level APIs hide framework internals. (b) TinyTorch students implement components like Adam, learning memory costs and update rules firsthand. (c) By Module 13, every import is student-built code, and transformers train on infrastructure they fully understand.}
\label{fig:code-comparison}
\end{figure*}

Building systems knowledge alongside ML fundamentals presents three pedagogical challenges: teaching systems thinking early without overwhelming beginners (\Cref{sec:systems}), managing cognitive load when teaching both algorithms and implementation (\Cref{sec:progressive}), and validating student understanding through concrete milestones (\Cref{subsec:milestones}). TinyTorch addresses these through curriculum design inspired by compiler courses~\citep{aho2006compilers}: students build a complete system incrementally, with each module adding functionality while maintaining a working implementation. This progression follows a deliberate dependency chain: tensors (Module 01) enable activations (02) and layers (03), which feed into dataloader (05) and autograd (06), powering optimizers (07) and training (08). Each completed module becomes immediately usable: after Module 03, students build neural networks; after Module 06, automatic differentiation enables training; after Module 13, transformers support language modeling. This structure enables students to gradually construct mental models while seeing immediate results.

TinyTorch is meant for students transitioning from framework \emph{users} to framework \emph{engineers}: those who have completed introductory ML courses (e.g., CS229, fast.ai) and want to understand PyTorch internals, those planning ML systems research or infrastructure careers, or practitioners debugging production deployment issues. The curriculum assumes NumPy proficiency and basic neural network familiarity but teaches framework architecture from first principles. Students needing immediate GPU/distributed training skills are better served by PyTorch tutorials; those preferring project-based application building will find high-level frameworks more appropriate. The 20-module structure supports flexible pacing: intensive completion, semester integration (parallel with lectures), or independent professional development.

This paper makes three contributions, each addressing the systems imperative the Bitter Lesson reveals:

\begin{enumerate}[leftmargin=*, itemsep=0.5em]
\item \textbf{Build-to-Validate Curriculum} (\Cref{sec:curriculum}): A 20-module learning path where students validate their implementations by recreating historical ML milestones, from Rosenblatt's Perceptron (1958) to modern transformers, using exclusively their own code. This provides concrete correctness criteria and grounds abstract concepts in tangible achievements.

\item \textbf{Progressive Disclosure of Complexity} (\Cref{sec:progressive}): A scaffolding technique that reveals \texttt{Tensor} internals gradually while maintaining a unified mental model. Gradient tracking infrastructure exists from Module 01 but activates only in Module 06, preventing premature cognitive overload while enabling seamless backpropagation later.

\item \textbf{Systems from Day One} (\Cref{sec:systems}): Memory profiling, computational complexity, and performance analysis are embedded starting in Module 01, not deferred to advanced topics. Students discover that Adam requires $2\times$ optimizer state memory by implementing it, not by reading documentation.
\end{enumerate}

\textbf{Paper Organization.} Before presenting TinyTorch's design, we position the contributions relative to existing educational frameworks and learning theories (\Cref{sec:related}). We then present the curriculum architecture (\Cref{sec:curriculum}), the progressive disclosure pattern that enables cognitive load management (\Cref{sec:progressive}), and systems-first integration throughout modules (\Cref{sec:systems}). Finally, we discuss deployment, limitations, empirical validation plans, and implications for ML education (\Cref{sec:deployment,sec:discussion,sec:conclusion}).


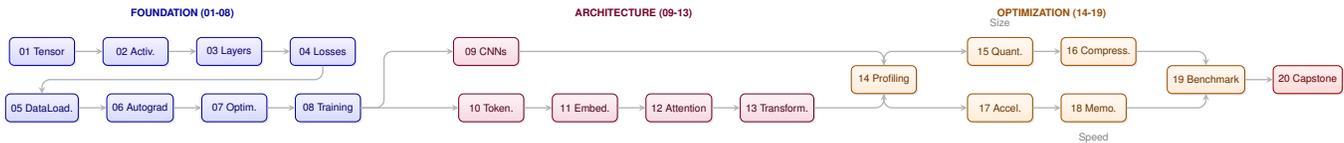
\begin{figure*}[t]
    \centering
    \resizebox{\textwidth}{!}{%
    \begin{tikzpicture}[
        node distance=0.4cm and 0.6cm,
        every node/.style={font=\scriptsize\sffamily},
        foundation/.style={
            rectangle, 
            draw=blue!60!black, 
            top color=blue!5, 
            bottom color=blue!15, 
            text=blue!40!black,
            minimum width=1.4cm, 
            minimum height=0.6cm, 
            rounded corners=3pt,
            drop shadow={opacity=0.2, shadow xshift=1pt, shadow yshift=-1pt}
        },
        architecture/.style={
            rectangle, 
            draw=purple!60!black, 
            top color=purple!5, 
            bottom color=purple!15, 
            text=purple!40!black,
            minimum width=1.4cm, 
            minimum height=0.6cm, 
            rounded corners=3pt,
            drop shadow={opacity=0.2, shadow xshift=1pt, shadow yshift=-1pt}
        },
        optimization/.style={
            rectangle, 
            draw=orange!60!black, 
            top color=orange!5, 
            bottom color=orange!15, 
            text=orange!40!black,
            minimum width=1.4cm, 
            minimum height=0.6cm, 
            rounded corners=3pt,
            drop shadow={opacity=0.2, shadow xshift=1pt, shadow yshift=-1pt}
        },
        capstone/.style={
            rectangle, 
            draw=red!60!black, 
            top color=red!5, 
            bottom color=red!15, 
            text=red!40!black,
            minimum width=1.4cm, 
            minimum height=0.6cm, 
            rounded corners=3pt,
            drop shadow={opacity=0.2, shadow xshift=1pt, shadow yshift=-1pt}
        },
        arr/.style={->, >=stealth, thick, gray!60},
        tier/.style={draw=gray!40, dashed, rounded corners=5pt, inner sep=8pt}
    ]
    
    \node[foundation] (T) {01 Tensor};
    \node[foundation, right=of T] (A) {02 Activ.};
    \node[foundation, right=of A] (L) {03 Layers};
    \node[foundation, right=of L] (Loss) {04 Losses};
    
    \node[foundation, below=0.6cm of T] (Data) {05 DataLoad.};
    \node[foundation, right=of Data] (Auto) {06 Autograd};
    \node[foundation, right=of Auto] (Opt) {07 Optim.};
    \node[foundation, right=of Opt] (Train) {08 Training};
    
    \draw[arr] (T) -- (A);
    \draw[arr] (A) -- (L);
    \draw[arr] (L) -- (Loss);
    \draw[arr, rounded corners=5pt] (Loss.south) -- ++(0,-0.3) -| (Data.north);
    \draw[arr] (Data) -- (Auto);
    \draw[arr] (Auto) -- (Opt);
    \draw[arr] (Opt) -- (Train);
    
    
        \node[architecture, right=of Loss, xshift=1.5cm] (Spatial) {09 CNNs};
        
        \node[architecture, right=of Train, xshift=1.5cm] (Tok) {10 Token.};
        \node[architecture, right=of Tok] (Emb) {11 Embed.};
        \node[architecture, right=of Emb] (Att) {12 Attention};
        \node[architecture, right=of Att] (Trans) {13 Transform.};
    
        \draw[arr, rounded corners=5pt] (Train.east) -- ++(0.5,0) |- (Spatial.west);
        \draw[arr] (Train) -- (Tok);
    
        \draw[arr] (Tok) -- (Emb);
        \draw[arr] (Emb) -- (Att);
        \draw[arr] (Att) -- (Trans);
    
        
    \path ($(Trans.east) + (1.5, 0)$) coordinate (ProfX);
    \path ($(Spatial)!0.5!(Trans)$) coordinate (ProfY);
    \node[optimization] (Prof) at (ProfX |- ProfY) {14 Profiling};
    
        \draw[arr, rounded corners=5pt] (Spatial.east) -| (Prof.north);
        \draw[arr, rounded corners=5pt] (Trans.east) -| (Prof.south);
    
        \node[optimization] (Quant) at (Prof |- Spatial) [xshift=2.5cm] {15 Quant.};
        \node[optimization, right=of Quant] (Comp) {16 Compress.};
        
        \node[optimization] (Accel) at (Prof |- Trans) [xshift=2.5cm] {17 Accel.};
        \node[optimization, right=of Accel] (Memo) {18 Memo.};
    
        \draw[arr, rounded corners=5pt] (Prof.north) |- (Quant.west);
        \draw[arr, rounded corners=5pt] (Prof.south) |- (Accel.west);
    
        \draw[arr] (Quant) -- (Comp);
        \draw[arr] (Accel) -- (Memo);
    
    \path ($(Comp.east) + (1.5, 0)$) coordinate (BenchX);
    \path ($(Comp)!0.5!(Memo)$) coordinate (BenchY);
    \node[optimization] (Bench) at (BenchX |- BenchY) {19 Benchmark};
    
        \draw[arr, rounded corners=5pt] (Comp.east) -| (Bench.north);
        \draw[arr, rounded corners=5pt] (Memo.east) -| (Bench.south);
    
    \node[capstone, right=of Bench] (Cap) {20 Capstone};
    \draw[arr] (Bench) -- (Cap);

    
    \node[font=\scriptsize\bfseries\sffamily, blue!60!black] at ($(T.north)!0.5!(Loss.north) + (0, 0.5)$) {FOUNDATION (01-08)};

    \path ($(Spatial.west)!0.5!(Trans.east)$) coordinate (ArchCenterX);
    \node[font=\scriptsize\bfseries\sffamily, purple!60!black] at (ArchCenterX |- Spatial.north) [yshift=0.5cm] {ARCHITECTURE (09-13)};

    \path ($(Quant.west)!0.5!(Comp.east)$) coordinate (OptCenterX);
    \node[font=\scriptsize\bfseries\sffamily, orange!60!black] at (OptCenterX |- Quant.north) [yshift=0.5cm] {OPTIMIZATION (14-19)};
    
        \node[above=0.1cm of Quant, font=\scriptsize\sffamily, gray] {Size};
        \node[below=0.1cm of Memo, font=\scriptsize\sffamily, gray] {Speed};
    
    \end{tikzpicture}%
    }
    \caption{\textbf{Module Dependency Graph.} TinyTorch's 20 modules form a directed acyclic graph with two architectural paths. Foundation modules (blue, M01--08) build core infrastructure sequentially, culminating in Training (M08). From Training, two paths branch: the \emph{Vision} path (M09) builds CNNs for spatial processing; the \emph{Language} path (M10--13) builds tokenization through transformers. Both paths converge at Profiling (M14), then branch into parallel optimization tracks---\emph{Model-level} (quantization, compression) and \emph{Runtime} (acceleration, memoization)---before final convergence at Benchmarking (M19) and Capstone (M20).}
    \label{fig:module-flow}
    \end{figure*}

\section{Related Work}
\label{sec:related}

TinyTorch builds upon decades of work in CS education research and recent innovations in ML framework pedagogy. We first establish the pedagogical tradition that grounds our approach, then position TinyTorch relative to existing ML educational frameworks.

\subsection{Pedagogical Precedents in Systems Edu}

TinyTorch joins a long pedagogical tradition in systems education: teaching complex systems by having students build simplified, transparent implementations. This approach emerged because production systems optimize for performance rather than comprehension, making their internals inaccessible to learners.

MINIX~\citep{tanenbaum1987minix} exemplifies this pattern. Tanenbaum created a teaching operating system despite Unix's existence because production kernels are too complex for students to hold in their heads. Students who implement a microkernel understand scheduler mechanics, memory management, and process isolation in ways that remain opaque to Linux users. MINIX shaped generations of systems engineers and famously inspired Linux itself. The parallel to TinyTorch is direct: PyTorch is Unix; TinyTorch is MINIX.

Compiler education follows identical logic. Despite LLVM and GCC representing decades of engineering excellence, compiler courses continue teaching from-scratch implementation~\citep{aho2006compilers,appel2004tiger}. Students build lexical analyzers, construct abstract syntax trees, implement type checkers, and generate code. The Tiger compiler~\citep{appel2004tiger} became canonical precisely because industrial compilers are too vast for pedagogical dissection. Students who build compilers understand optimization passes and code generation in ways that framework users cannot.

Operating systems education institutionalized this approach through purpose-built teaching systems. Nachos~\citep{christopher1993nachos} at Berkeley and Pintos~\citep{pfaff2004pintos} at Stanford provided hackable kernels where students implement threads, schedulers, virtual memory, and file systems. More recently, xv6~\citep{kaashoek2023xv6} at MIT demonstrated that even established teaching systems benefit from architectural simplification: transitioning from x86 to RISC-V stripped away historical complexity to expose cleaner abstractions. These systems never aimed to replace Linux; they aimed to reveal what Linux hides. The pedagogical insight transfers directly: TinyTorch is to PyTorch what xv6 is to Unix.

Perhaps most fundamentally, SICP~\citep{abelson1996sicp} taught programming through building interpreters and evaluators from scratch, despite real languages already existing. The core philosophy: to understand a system, build one. This principle, validated across five decades of CS education, grounds TinyTorch's design.

These canonical precedents share common characteristics: (1) production systems grew too complex to learn from directly, (2) lightweight reconstructions provided genuine insight unavailable through usage alone, (3) educational versions outlived production systems in pedagogical value, and (4) they trained systems thinkers rather than mere users. TinyTorch applies this proven pattern to ML frameworks at the precise moment when PyTorch and TensorFlow have matured beyond pedagogical accessibility, and when industry desperately needs engineers who understand framework internals.

\subsection{Educational ML Frameworks}

Educational frameworks teaching ML internals occupy different points in the scope-simplicity tradeoff space. \textbf{micrograd}~\citep{karpathy2022micrograd} demonstrates autograd mechanics elegantly in approximately 200 lines of scalar-valued Python, making backpropagation transparent through decomposition into elementary operations. Its pedagogical clarity comes from intentional minimalism: scalar operations only, no tensor abstraction, focused solely on automatic differentiation fundamentals. This design illuminates gradient mechanics but necessarily omits systems concerns (memory profiling, computational complexity, production patterns) and modern architectures.

\textbf{MiniTorch}~\citep{schneider2020minitorch} extends beyond autograd to tensor operations, neural network modules, and optional GPU programming, originating from Cornell Tech's Machine Learning Engineering course. The curriculum progresses from foundational autodifferentiation through deep learning with assessment infrastructure (unit tests, visualization tools). While MiniTorch includes an optional GPU module exploring parallel programming concepts and covers efficiency considerations throughout, the core curriculum emphasizes mathematical rigor: students work through detailed exercises building tensor abstractions from first principles. TinyTorch differs through systems-first emphasis (memory profiling and complexity analysis embedded from Module 01), production-inspired package organization, and three integration models supporting diverse deployment contexts.

\textbf{tinygrad}~\citep{hotz2023tinygrad} positions itself between micrograd's simplicity and PyTorch's production capabilities, providing a complete framework (tensor library, IR, compiler, JIT) that emphasizes hackability and transparency. Unlike opaque production frameworks, tinygrad makes ``the entire compiler and IR visible,'' enabling students to understand deep learning compilation internals. While pedagogically valuable through its inspectable design, tinygrad assumes significant background: students must navigate compiler concepts, multiple hardware backends, and production-level architecture without scaffolded progression or automated assessment infrastructure.

\textbf{Stanford CS231n}~\citep{johnson2016cs231n}, \textbf{CMU Deep Learning Systems (CS 10-414)}~\citep{chen2022dlsyscourse}, and \textbf{Harvard TinyML}~\citep{banbury2021widening} represent university courses that include implementation components with different systems emphases. CS231n's assignments involve NumPy implementations of CNNs, backpropagation, and optimization algorithms, providing hands-on experience with neural network internals. However, assignments are isolated exercises rather than cumulative framework construction, and systems concerns (memory profiling, complexity analysis) are not embedded from the start. CMU's DL Systems course explicitly targets ML systems engineering, covering automatic differentiation, GPU programming, distributed training, and deployment: representing the production systems knowledge TinyTorch provides conceptual foundations for. Harvard's TinyML Professional Certificate focuses on deploying ML to resource-constrained embedded devices (microcontrollers with KB-scale memory), teaching TensorFlow Lite for Microcontrollers through Arduino-based projects. While TinyML emphasizes hardware constraints and embedded deployment (achieving systems thinking through resource limitations), TinyTorch focuses on framework internals and algorithmic understanding (achieving systems thinking through implementation transparency). TinyML students learn \emph{how to optimize for} hardware constraints; TinyTorch students learn \emph{why frameworks work} internally. These approaches complement rather than compete: TinyML prepares students for edge deployment, TinyTorch for framework engineering and infrastructure development.

\textbf{Dive into Deep Learning (d2l.ai)}~\citep{zhang2021dive} and \textbf{fast.ai}~\citep{howard2020fastai} represent comprehensive ML education but with different pedagogical emphases than framework construction. d2l.ai provides interactive implementations across multiple frameworks (PyTorch, JAX, TensorFlow, MXNet/NumPy) through executable notebooks, teaching algorithmic foundations alongside practical coding. The NumPy implementation track includes from-scratch implementations of key algorithms, though these are presented as educational demonstrations rather than components of a cumulative framework students build. With widespread adoption across hundreds of universities globally, it excels at algorithmic understanding through framework usage. fast.ai's distinctive top-down pedagogy starts with practical applications before foundations, using layered APIs that provide high-level abstractions while enabling deeper exploration through PyTorch. Both resources assume cloud computing access (AWS, Google Colab, SageMaker) for GPU-based training, though provide various deployment options.

\subsection{Positioning and Unique Contributions}

TinyTorch occupies a distinct pedagogical niche through its \textbf{bottom-up, systems-first approach}. Unlike top-down pedagogies (fast.ai: start with applications, descend to details) or algorithm-focused curricula (d2l.ai: master theory through framework usage), TinyTorch employs bottom-up framework construction: students build core abstractions first (tensors, autograd, layers), then compose them into architectures (CNNs, transformers), finally optimizing for production constraints (quantization, compression). This grounds systems thinking in direct implementation rather than abstract instruction. The curriculum serves students post-introductory-ML (ready to transition from framework users to engineers), pre-systems-research (providing foundations before production ML courses like CMU DL Systems), and complementary to algorithm courses (adding systems awareness to mathematical foundations).

\Cref{tab:framework-comparison} positions TinyTorch relative to both educational frameworks (micrograd, MiniTorch, tinygrad) and production frameworks (PyTorch, TensorFlow), clarifying that TinyTorch serves as pedagogical bridge between understanding frameworks and using them professionally.

\begin{table*}[tp]
\centering
\caption{Framework comparison positions TinyTorch's pedagogical role. Educational frameworks (micrograd, MiniTorch, tinygrad) prioritize learning over production use. Production frameworks (PyTorch, TensorFlow) prioritize performance and scalability. TinyTorch bridges both: students learn framework internals through implementation, then transfer that knowledge to production frameworks with deeper systems understanding.}
\label{tab:framework-comparison}
\small
\renewcommand{\arraystretch}{1.4}
\setlength{\tabcolsep}{6pt}
\begin{tabularx}{0.98\textwidth}{@{}>{\raggedright\arraybackslash}p{2.0cm}>{\raggedright\arraybackslash}p{2.0cm}>{\raggedright\arraybackslash}X>{\raggedright\arraybackslash}X>{\raggedright\arraybackslash}X@{}}
\toprule
\textbf{Framework} & \textbf{Purpose} & \textbf{Scope} & \textbf{Systems Focus} & \textbf{Target Outcome} \\
\midrule
\multicolumn{5}{@{}l}{\textbf{Educational Frameworks}} \\
\addlinespace[2pt]
micrograd & Teach autograd & Autograd only (scalar) & Minimal & Understand backprop \\
MiniTorch & Teach ML math & Tensors + autograd + optional GPU & Math foundations & Build from first principles \\
tinygrad & Inspectable production & Complete (compiler, IR, JIT) & Advanced (compiler) & Understand compilation \\
\addlinespace[1pt]
\textbf{TinyTorch} & \textbf{Teach ML systems} & \textbf{Complete (tensors $\rightarrow$ transformers $\rightarrow$ optimization)} & \textbf{Embedded from Module 01} & \textbf{Framework engineers} \\
\addlinespace[2pt]
\midrule
\multicolumn{5}{@{}l}{\textbf{Production Frameworks}} \\
\addlinespace[2pt]
PyTorch & Production ML & Complete (GPU, distributed, deployment) & Advanced (implicit) & Train models efficiently \\
TensorFlow & Production ML & Complete (GPU, distributed, deployment, mobile) & Advanced (implicit) & Deploy at scale \\
\bottomrule
\end{tabularx}
\end{table*}

TinyTorch differs from educational frameworks through systems-first integration and from production frameworks through pedagogical transparency. Key distinctions: versus micrograd, complete framework scope with assessment infrastructure; versus MiniTorch, systems-first emphasis with unified API evolution; versus tinygrad, scaffolded progression without compiler prerequisites; versus d2l.ai and fast.ai, bottom-up framework construction rather than algorithm mastery or top-down application focus.

Empirical validation of learning outcomes remains future work (\Cref{sec:discussion}), but design grounding in established learning theory provides theoretical justification for pedagogical choices.

\section{TinyTorch Architecture}
\label{sec:curriculum}

This section presents TinyTorch's curriculum architecture. The design is grounded in established learning theory: constructionism~\citep{papert1980mindstorms} (learning through building artifacts), cognitive apprenticeship~\citep{collins1989cognitive} (making expert thinking visible), productive failure~\citep{kapur2008productive} (struggling before instruction), and threshold concepts~\citep{meyer2003threshold} (transformative ideas requiring mastery). These principles shaped the competency framework and module progression that follow.

\subsection{The ML Systems Competency Matrix}
\label{subsec:competency-matrix}

The learning theories above informed a concrete competency framework that operationalizes pedagogical goals into measurable outcomes. What must a student master to understand ML systems fundamentals? We decompose this into 8 knowledge areas progressing from foundational (tensors) to production (deployment), crossed with 5 capability levels progressing from conceptual understanding to optimization mastery (\Cref{tab:competency-matrix}). This matrix preceded module design: we first identified the 40 competency cells, then engineered modules to systematically address them.

\begin{table*}[t]
\centering
\caption{\textbf{ML Systems Competency Matrix (Single Node).} The 40 cells (8 knowledge areas $\times$ 5 capability levels) characterize the foundational competencies that AI engineers require. Rows progress from foundational concepts (tensors, operations) through learning mechanics (graphs, optimization) to production concerns (efficiency, deployment). Columns progress from conceptual understanding (Knows) through quantitative reasoning (Measures) and implementation (Implements) to diagnostic skill (Debugs) and performance engineering (Optimizes). Distributed systems concepts (gradient synchronization, model parallelism, communication overhead) represent a natural subsequent learning objective building on these foundations. This matrix guided TinyTorch's curriculum design. \Cref{subsec:coverage} maps the 20 modules to this framework.}
\label{tab:competency-matrix}
\small
\renewcommand{\arraystretch}{1.4}
\setlength{\tabcolsep}{6pt}
\begin{tabularx}{0.98\textwidth}{@{}>{\raggedright\arraybackslash}p{2.4cm}>{\raggedright\arraybackslash}X>{\raggedright\arraybackslash}X>{\raggedright\arraybackslash}X>{\raggedright\arraybackslash}X>{\raggedright\arraybackslash}X@{}}
\toprule
\textbf{Knowledge Area} & \textbf{Knows} & \textbf{Measures} & \textbf{Implements} & \textbf{Debugs} & \textbf{Optimizes} \\
\midrule
Tensors \& Memory & Layouts, strides, dtypes & Bytes, peak allocation & Tensor class, broadcasting & OOM, memory leaks & In-place ops, pooling \\
\addlinespace[2pt]
Operations \& Compute & Broadcast rules, semantics & FLOPs, arithmetic intensity & MatMul, Conv2d, activations & Shape mismatch, instability & Vectorization, fusion \\
\addlinespace[2pt]
Graphs \& Differentiation & Computation graphs, memory lifecycle & Gradient memory, node count & Autograd, backward() & Vanishing/ exploding grads & Checkpointing \\
\addlinespace[2pt]
Optimization \& Training & Optimizer state, update costs & State size, convergence rate & Optimizers, training loop & Non-convergence, NaN & LR scheduling, clipping \\
\addlinespace[2pt]
Architectures \& Models & Parameter scaling, compute patterns & Parameters, receptive field & Layers, blocks, full models & Accuracy plateau, overfit & Pruning, distillation \\
\addlinespace[2pt]
Pipelines \& Data & Batching, shuffling, prefetch & Throughput, CPU utilization & Dataset, DataLoader & Data bottleneck, stalls & Parallel loading, caching \\
\addlinespace[2pt]
Efficiency \& Compression & Quantization, pruning tradeoffs & Compression ratio, accuracy delta & INT8, magnitude pruning & Accuracy degradation & Calibration, sparsity \\
\addlinespace[2pt]
Deployment \& Serving & Batch vs streaming, caching & Latency p50/p99, throughput & KV cache, model export & Timeout, memory bloat & Dynamic batching \\
\bottomrule
\end{tabularx}
\end{table*}

The matrix embodies two design principles. First, \textbf{rows are progressive}: each knowledge area builds on those above. Students cannot understand optimization without first understanding gradients; deployment concerns assume architecture mastery. This ordering directly informs TinyTorch's module sequencing. Second, \textbf{columns are progressive}: capability levels build left-to-right. Students must understand concepts (Knows) before quantifying costs (Measures), must measure before implementing (Implements), must implement before diagnosing failures (Debugs), and must debug before systematically improving (Optimizes). This progression shapes each module's internal structure through the Build $\rightarrow$ Use $\rightarrow$ Reflect cycle (\Cref{subsec:module-pedagogy}).

Not every cell requires equal depth. An educational framework prioritizes core cells (Tensors through Training, columns 1--4) over production-specific concerns (Deployment-Optimizes). \Cref{subsec:coverage} maps our actual coverage, including intentional gaps where production infrastructure exceeds accessibility-first scope.

\subsection{The 3-Tier Learning Journey + Capstone}

TinyTorch organizes modules into three progressive tiers plus a capstone competition (\Cref{tab:objectives}). Students cannot skip tiers: architectures require foundation mastery, optimization demands training system understanding. The tiers mirror ML systems engineering practice: foundation (core ML mechanics), architectures (domain-specific models), optimization (production deployment), culminating in the Capstone (competitive systems engineering).

\begin{table*}[p]
\centering
\caption{\textbf{Module Progression.} Each module teaches both ``what'' (ML technique) and ``how much'' (memory/compute costs). Foundation tier (M01--08) establishes core operations with explicit resource tracking. Architecture tier (M09--13) applies these foundations to CNNs and transformers. Optimization tier (M14--19) adds production concerns: profiling, quantization, deployment. This dual-concept approach ensures students never learn algorithms without understanding their systems implications.}
\label{tab:objectives}
\small
\renewcommand{\arraystretch}{1.4}
\setlength{\tabcolsep}{6pt}
\begin{tabularx}{0.98\textwidth}{@{}cl>{\raggedright\arraybackslash}p{2.2cm}>{\raggedright\arraybackslash}X>{\raggedright\arraybackslash}X@{}}
\toprule
\textbf{Mod} & \textbf{Tier} & \textbf{Module Name} & \textbf{ML Concept} & \textbf{Systems Concept} \\
\midrule
\multicolumn{5}{@{}l}{\textbf{Foundation Tier (01--08)}} \\
\addlinespace[2pt]
01 & Fnd & Tensor & Multidimensional arrays, broadcasting & Memory footprint (nbytes), dtype sizes, contiguous layout \\
02 & Fnd & Activations & ReLU, Sigmoid, Tanh, GELU, Softmax & Numerical stability (exp overflow), vectorization \\
03 & Fnd & Layers & Linear, Xavier initialization & Parameter vs activation memory, weight layout \\
04 & Fnd & Losses & Cross-entropy, MSE, log-sum-exp trick & Numerical stability (log(0)), gradient magnitude \\
05 & Fnd & DataLoader & Dataset abstraction, batching, shuffling & Iterator protocol, batch collation overhead \\
06 & Fnd & Autograd & Computational graphs, chain rule, backprop & Gradient memory (2$\times$ momentum, 3$\times$ Adam) \\
07 & Fnd & Optimizers & SGD, Momentum, Adam, AdamW & Optimizer state memory, in-place updates \\
08 & Fnd & Training & Cosine scheduling, gradient clipping & Peak memory lifecycle, checkpoint tradeoffs \\
\addlinespace[2pt]
\midrule
\multicolumn{5}{@{}l}{\textbf{Architecture Tier (09--13)}} \\
\addlinespace[2pt]
09 & Arch & Convolutions & Conv2d, pooling, padding, stride & im2col expansion, 7-loop $O(B \!\times\! C \!\times\! H \!\times\! W \!\times\! K^2)$ \\
10 & Arch & Tokenization & BPE, vocabulary, special tokens & Vocab size$\leftrightarrow$sequence length tradeoff \\
11 & Arch & Embeddings & Token + positional (sinusoidal/learned) & Sparse gradient updates, embedding table memory \\
12 & Arch & Attention & Scaled dot-product, causal masking & $O(N^2)$ memory, attention score materialization \\
13 & Arch & Transformers & Multi-head attention, LayerNorm, MLP & KV cache sizing, per-layer memory profile \\
\addlinespace[2pt]
\midrule
\multicolumn{5}{@{}l}{\textbf{Optimization Tier (14--19)}} \\
\addlinespace[2pt]
14 & Opt & Profiling & Time/memory/FLOPs measurement & Bottleneck identification, measurement overhead \\
15 & Opt & Quantization & INT8, scale/zero-point calibration & 4$\times$ compression, quantization error propagation \\
16 & Opt & Compression & Magnitude pruning, knowledge distillation & Sparsity patterns, teacher-student memory \\
17 & Opt & Acceleration & Vectorization, memory access patterns & Cache locality, SIMD utilization \\
18 & Opt & Memoization & KV-cache for autoregressive generation & $O(n^2)$$\rightarrow$$O(n)$ caching, memory-compute tradeoff \\
19 & Opt & Benchmarking & Statistical comparison, multiple runs & Confidence intervals, warm-up protocols \\
\addlinespace[2pt]
\midrule
\multicolumn{5}{@{}l}{\textbf{Capstone (20)}} \\
\addlinespace[2pt]
20 & Cap & Capstone & End-to-end optimized system, benchmark submission & MLPerf-style metrics, JSON validation, leaderboard integration \\
\bottomrule
\end{tabularx}
\end{table*}

\begin{lstlisting}[caption={\textbf{Memory Profiling.} Tensor implementation from Module 01 with explicit memory tracking.},label={lst:tensor-memory},float=t]
class Tensor:
  def __init__(self, data):
    self.data = np.array(data, dtype=np.float32)
    self.shape = self.data.shape

  def memory_footprint(self):
    """Calculate exact memory in bytes"""
    return self.data.nbytes

  def __matmul__(self, other):
    if self.shape[-1] != other.shape[0]:
      raise ValueError(
        f"Shape mismatch: {self.shape} @ {other.shape}"
      )
    return Tensor(self.data @ other.data)
\end{lstlisting}

\textbf{Tier 1: Foundation (Modules 01--08).}
Students build the mathematical core enabling neural networks to learn, following a deliberate \emph{Forward Pass $\rightarrow$ Learning Infrastructure $\rightarrow$ Training} progression. Modules 01--04 construct forward pass components in the order data flows: tensors (data structure), activations (non-linearity), layers (parameterized transformations), and losses (objective functions). Systems thinking begins immediately: Module 01 introduces \texttt{memory\_footprint()} before matrix multiplication (\Cref{lst:tensor-memory}), making memory a first-class concept. Modules 05--07 build learning infrastructure: data loading (Module 05) provides efficient batching, then automatic differentiation (Module 06) enables gradient computation through progressive disclosure (\Cref{sec:progressive}), and optimizers (Module 07) use those gradients for parameter updates. Students discover Adam's 3$\times$ optimizer-related memory overhead (gradients plus two state buffers) through direct measurement (\Cref{sec:systems}). The training loop (Module 08) integrates all components. This order is the minimal dependency chain: you cannot build optimizers without autograd (no gradients), cannot build autograd without losses (nothing to differentiate), cannot build losses without layers (no predictions). By tier completion, students recreate three historical milestones: \citet{rosenblatt1958perceptron}'s Perceptron, Minsky and Papert's XOR solution, and \citet{rumelhart1986learning}'s backpropagation targeting 95\%+ on MNIST.

\textbf{Tier 2: Architectures (Modules 09--13).}
Students apply foundation knowledge to modern architectures, with the tier branching into parallel \emph{Vision} and \emph{Language} tracks. This bifurcation reflects domain-specific requirements: vision processes spatial grids (images), while language processes variable-length sequences (text). Both tracks build on the DataLoader patterns from Module 05 and training infrastructure from Module 08. TinyTorch ships with two custom educational datasets: \textbf{TinyDigits} ($\sim$1,000 grayscale handwritten digits) and \textbf{TinyTalks} ($\sim$350 conversational Q\&A pairs). These datasets are deliberately small and offline-first: they require no network connectivity during training, consume minimal storage ($<$50MB combined), and train in minutes on CPU-only hardware. This design ensures accessibility for students in regions with limited internet infrastructure, institutional computer labs with restricted network access, and developing countries where cloud-based datasets create barriers to ML education.

The tier branches into two paths, each following the principle of building components in the order they compose. \textbf{Vision} implements Conv2d with seven explicit nested loops making $O(C_{out} \times H \times W \times C_{in} \times K^2)$ complexity visible before optimization. Students discover weight sharing's dramatic efficiency through direct comparison: Conv2d(3$\rightarrow$32, kernel=3) requires 896 parameters while an equivalent dense layer needs 98,336 parameters (3072 input features $\times$ 32 outputs + 32 bias terms), a 109$\times$ reduction demonstrating how inductive biases enable CNNs to learn spatial patterns without brute-force parameterization. Students validate their implementations by training CNNs targeting 65--75\% CIFAR-10 accuracy~\citep{krizhevsky2009cifar,lecun1998gradient}.

\textbf{Language} progresses through tokenization, embeddings, attention, and complete transformers~\citep{vaswani2017attention}, with each step transforming the data representation in sequence (text $\rightarrow$ token IDs $\rightarrow$ embeddings $\rightarrow$ contextualized representations). This order is non-negotiable: you cannot compute attention without embeddings (attention needs continuous vectors), cannot create embeddings without tokenization (embedding lookup needs integer indices), cannot tokenize without text (input must exist). Module 10 (Tokenization) teaches a fundamental NLP systems trade-off: vocabulary size controls model parameters (embedding matrix rows $\times$ dimensions), while sequence length determines transformer computation ($O(n^2)$ attention complexity). Students discover why later GPT models increased vocabulary from 50K tokens (GPT-2/GPT-3) to 100K tokens (GPT-3.5/GPT-4): not for better language understanding, but to reduce sequence lengths for long documents, trading parameter memory for computational efficiency. Students experience quadratic scaling through direct measurement, validating their transformer implementations through text generation on TinyTalks.

\textbf{Tier 3: Optimization (Modules 14--19).}
Students transition from ``models that train'' to ``systems that deploy.'' The optimization tier follows a deliberate pedagogical structure: \emph{Measure $\rightarrow$ Model-Level $\rightarrow$ Runtime $\rightarrow$ Validate}. Profiling (14) teaches measuring time, memory, and FLOPs (floating-point operations), introducing Amdahl's Law: optimizing 70\% of runtime by 2$\times$ yields only 1.53$\times$ overall speedup because the remaining 30\% becomes the new bottleneck. This teaches that optimization is iterative and measurement-driven.

The tier then divides into two optimization categories. \textbf{Model-level optimizations} (Modules 15--16) change the model itself: Quantization (15) achieves 4$\times$ compression (FP32$\rightarrow$INT8) with 1--2\% accuracy cost, while Compression (16) applies pruning and distillation for 10$\times$ shrinkage. These techniques permanently modify model weights and architecture.

\textbf{Runtime optimizations} (Modules 17--18) change how execution happens without modifying model weights. Acceleration (17) teaches general-purpose optimization: vectorization exploits SIMD instructions for 10--100$\times$ convolution speedups, memory access pattern optimization improves cache locality, and kernel fusion eliminates intermediate memory traffic. These techniques apply to any numerical computation. Memoization (18) then applies domain-specific optimization to transformers through KV caching: students discover that naive autoregressive generation recomputes attention keys and values at every step, so generating 100 tokens requires 5,050 redundant computations (1+2+...+100). By caching these values, students transform $O(n^2)$ generation into $O(n)$, achieving 10--100$\times$ speedup and understanding why this optimization is essential in systems like ChatGPT and Claude for economically viable inference.

Benchmarking (19) teaches statistical rigor in performance measurement: students learn that single measurements are meaningless (performance varies 10--30\% across runs due to thermal throttling, OS noise, and cache state), implement confidence intervals and warmup protocols, and discover when a 5\% speedup is statistically significant versus noise.

\textbf{Capstone (Module 20).}
The capstone builds submission infrastructure enabling ML competitions and reproducible benchmarking. Inspired by MLPerf~\citep{mattson2020mlperf,reddi2020mlperf}, students implement standardized submission formats (JSON schemas with system metadata, normalized metrics, and improvement calculations), benchmark reporting classes, and complete optimization workflows. This infrastructure serves dual purposes: students show mastery by applying the complete optimization pipeline (Profile $\rightarrow$ Optimize $\rightarrow$ Benchmark $\rightarrow$ Submit) to their prior implementations, and the submission system enables future community challenges (classroom competitions, optimization contests, reproducible research comparisons). Students learn that ML engineering requires not just optimization techniques but shareable, reproducible infrastructure for validating and comparing results.

\subsection{Module Structure}
\label{subsec:module-pedagogy}

Each module follows a consistent \textbf{Build $\rightarrow$ Use $\rightarrow$ Reflect} pedagogical cycle that integrates implementation, application, and systems reasoning. This structure addresses multiple learning objectives: students construct working components (Build), validate integration with prior modules (Use), and develop systems thinking~\citep{meadows2008thinking} through analysis (Reflect).

\textbf{Build: Implementation with Explicit Dependencies.} Students implement components in Jupyter notebooks (\texttt{*.py}) with scaffolded guidance. Each module begins with a \emph{connection map}: a visual diagram showing which prior modules students must have completed (prerequisites), what the current module teaches (focus), and what capabilities become available after completion (unlocks). For example, Module 09 (Convolutions) shows prerequisites of Modules 01--08, focus on spatial operations, and unlocks CNN architectures for image classification. These maps make dependency relationships explicit, helping students understand where each module fits in the larger framework architecture. To maintain consistency across all 20 modules, Module 05 (DataLoader) serves as the canonical reference implementation that all modules follow, reducing maintenance burden and enabling consistent community contribution.

\textbf{Use: Integration Testing Beyond Unit Tests.} Assessment validates both isolated correctness and cross-module integration. Unit tests verify individual component behavior (``Does \texttt{Tensor.reshape()} produce correct output?''), while integration tests validate that components compose into working systems (``Can Module 06 Autograd compute gradients through Module 03 Linear layers?''). Integration tests are critical for TinyTorch's pedagogical model because students may pass Module 03 unit tests but fail when autograd activates in Module 06: their layer implementation doesn't properly propagate \texttt{requires\_grad} through operations or construct computational graphs correctly.

A common failure pattern illustrates this: students implement \texttt{Linear.forward()} that passes unit tests (correct output values), but gradients don't flow during backpropagation because they used NumPy operations directly instead of \texttt{Tensor} operations. When \texttt{x.requires\_grad=True} flows into their layer, the computational graph breaks. Students encounter errors like ``\texttt{AttributeError: 'numpy.ndarray' object has no attribute 'backward'}'' and must debug interface contracts: operations must preserve \texttt{Tensor} types to maintain gradient connectivity. This teaches \emph{interface design}: components must satisfy contracts enabling composition, not just produce correct outputs in isolation.

Module 09 (Convolutions) integration exemplifies this: convolution must work with Module 06's autograd (gradient flow through kernels), Module 07's optimizers (parameter updates), and Module 08's training loop (forward-backward cycles) simultaneously. Students discover that ``passing unit tests'' $\neq$ ``works in the system'' when their Conv2d produces correct outputs but crashes during \texttt{loss.backward()} because they forgot to track intermediate activations for gradient computation. This debugging mirrors professional ML engineering: isolated correctness is insufficient; system integration reveals interface failures.

\textbf{Reflect: Systems Analysis Questions.}
Each module concludes with systems reasoning prompts measuring conceptual understanding beyond syntactic correctness. Memory analysis questions ask students to calculate footprints (``A (256, 256) Conv2d layer with 64 input and 128 output channels requires how much memory?''). Complexity analysis prompts probe asymptotic understanding (``Why is attention $O(N^2)$? Demonstrate by doubling sequence length and measuring memory growth.''). Design trade-off questions assess engineering judgment (``Adam requires 2$\times$ optimizer state memory (momentum and variance) but converges faster than SGD. When is the 4$\times$ total training memory trade-off worth it?''). These open-ended questions assess transfer~\citep{perkins1992transfer}: can students apply learned concepts to novel scenarios not seen in exercises?

\subsection{Milestone Arcs}
\label{subsec:milestones}

Milestones are validation checkpoints where students recreate historical ML breakthroughs using exclusively their own TinyTorch implementations. Rather than toy exercises, milestones are real tasks: training Rosenblatt's 1958 Perceptron, solving Minsky's XOR problem, classifying CIFAR-10 images with CNNs, generating text with transformers. Each milestone requires importing only from \texttt{tinytorch.*}, where every layer, optimizer, and training loop is code the student wrote. Six milestones span ML history from 1958 to present, with each requiring progressively more modules from the growing framework.

\textbf{Why Milestones Matter.} Milestones serve dual pedagogical and validation purposes that differentiate TinyTorch from traditional programming assignments. First, pedagogical motivation through historical framing: Rather than ``implement this function,'' students ``recreate the breakthrough that proved Minsky wrong about neural networks,'' connecting implementation work to historically significant results. This instantiates Bruner's spiral curriculum~\citep{bruner1960process}: students train neural networks 6 times with increasing sophistication, each iteration deepening understanding through historical progression from 1958 (Perceptron) to present (production-optimized systems).

Second, implementation validation beyond unit tests: Milestones address what we call the ``implementation-example gap'': students can pass all unit tests but fail milestone tasks due to composition errors. This gap arises because unit tests validate isolated correctness (``Does \texttt{Linear.forward()} produce correct output?''), while milestones validate system composition (``Does the training loop properly orchestrate forward passes, loss computation, and backpropagation?''). Students who pass all Module 01--08 unit tests might still fail Milestone 3 (MLP Revival) if their components don't compose correctly into functional systems. This mirrors professional ML engineering: individual functions may work, but the system fails due to integration bugs. If student-implemented CNNs successfully classify natural images, convolution, pooling, and backpropagation all work correctly together; if transformers generate coherent text, attention mechanisms integrate properly. Milestone success is measured by achieving performance in the ballpark of historical benchmarks (CNNs with reasonable CIFAR-10 accuracy, transformers generating coherent text), not matching exact published accuracies. The goal is demonstrating implementations work correctly on real tasks, validating framework correctness.

\textbf{The Six Historical Milestones.} The curriculum includes six milestones spanning 1958 to present, each requiring progressively more components from the growing framework:

\begin{enumerate}
\item \textbf{1958 Perceptron} (after Module 04): Train Rosenblatt's original single-layer perceptron on linearly separable classification. Students import \texttt{from tinytorch import Tensor, Linear, Sigmoid}, their framework now supports single-layer networks.

\item \textbf{1969 XOR Solution} (after Module 08): Solve Minsky's ``impossible'' XOR problem with multi-layer perceptrons, proving critics wrong. Validates that autograd enables non-linear learning.

\item \textbf{1986 MLP Revival} (after Module 08): Handwritten digit recognition demonstrating backpropagation's power. Requires Modules 01--08 working together (tensor operations, activations, layers, losses, dataloader, autograd, optimizers, training). Students import \texttt{from tinytorch import SGD, CrossEntropyLoss}, their framework trains multi-layer networks end-to-end on MNIST digits.

\item \textbf{1998 CNN Revolution} (after Module 09): Image classification demonstrating convolutional architectures' advantage~\citep{krizhevsky2009cifar,lecun1998gradient}. Students import \texttt{from tinytorch import Conv2d, MaxPool2d}, training both MLP and CNN on CIFAR-10 to measure architectural improvements themselves through direct comparison.

\item \textbf{2017 Transformer Era} (after Module 13): Language generation with attention-based architecture. Validates that attention mechanisms, positional embeddings, and autoregressive sampling function correctly through coherent text generation.

\item \textbf{2018--Present: MLPerf Benchmarks} (after Module 20): Building submission infrastructure for reproducible benchmarking and future competitions. MLPerf~\citep{mattson2020mlperf,reddi2020mlperf} established industry-standard benchmarking in 2018, and systems engineering has only grown more critical since. Students apply the complete optimization pipeline (Profile $\rightarrow$ Optimize $\rightarrow$ Benchmark $\rightarrow$ Submit) to prior implementations, generating standardized results. This validates that students understand production ML engineering: not just optimization techniques, but shareable infrastructure for validating and comparing results. These are the systems skills that now determine who can actually deploy at scale.
\end{enumerate}

Each milestone: (1) recreates actual breakthroughs using exclusively student code, (2) uses \emph{only} TinyTorch implementations (no PyTorch/TensorFlow), (3) validates success through task-appropriate performance, and (4) demonstrates architectural comparisons showing why new approaches improved over predecessors.

\textbf{Validation:}
While milestones provide pedagogical motivation through historical framing, they simultaneously serve a technical validation purpose: demonstrating implementation correctness through real-world task performance. Success criteria for each milestone:

\begin{itemize}[leftmargin=*, itemsep=1pt, parsep=0pt]
    \item \textbf{Milestone 1 (1958 Perceptron)}: Solves linearly separable problems (e.g., 4-point OR/AND tasks), demonstrating basic gradient descent convergence.
    \item \textbf{Milestone 2 (1969 XOR Solution)}: Solves XOR classification, proving multi-layer networks handle non-linear problems that single layers cannot.
    \item \textbf{Milestone 3 (1986 MLP Revival)}: Achieves strong MNIST digit classification accuracy, validating backpropagation through all layers of deep networks.
    \item \textbf{Milestone 4 (1998 CNN Revolution)}: Achieves meaningful CIFAR-10 classification accuracy, showing convolutional feature extraction and spatial hierarchies work correctly.
    \item \textbf{Milestone 5 (2017 Transformer)}: Generates coherent multi-token text continuations on TinyTalks dataset, demonstrating functional attention mechanisms and autoregressive generation.
    \item \textbf{Milestone 6 (2018--Present: MLPerf Benchmarks)}: Generates valid benchmark submissions with reproducible metrics following the optimization pipeline, demonstrating mastery of systems engineering practices essential for production deployment.
\end{itemize}

Performance targets differ from published state-of-the-art due to pure-Python constraints (no GPU acceleration, simplified architectures). Correctness matters more than speed: if a student's CNN learns meaningful CIFAR-10 features, their convolution, pooling, and backpropagation implementations compose correctly into a functional vision system. This approach mirrors professional debugging where implementations prove correct by solving real tasks, not by passing synthetic unit tests alone.

\section{Progressive Disclosure}
\label{sec:progressive}

This section details how TinyTorch implements progressive disclosure: a pattern that manages cognitive load by enhancing \texttt{Tensor} capabilities gradually through monkey-patching while maintaining a unified mental model. Unlike approaches that introduce separate classes or require students to learn new APIs mid-curriculum, students work with a single \texttt{Tensor} class throughout, and its capabilities expand transparently as modules progress.

\subsection{Pattern Implementation}

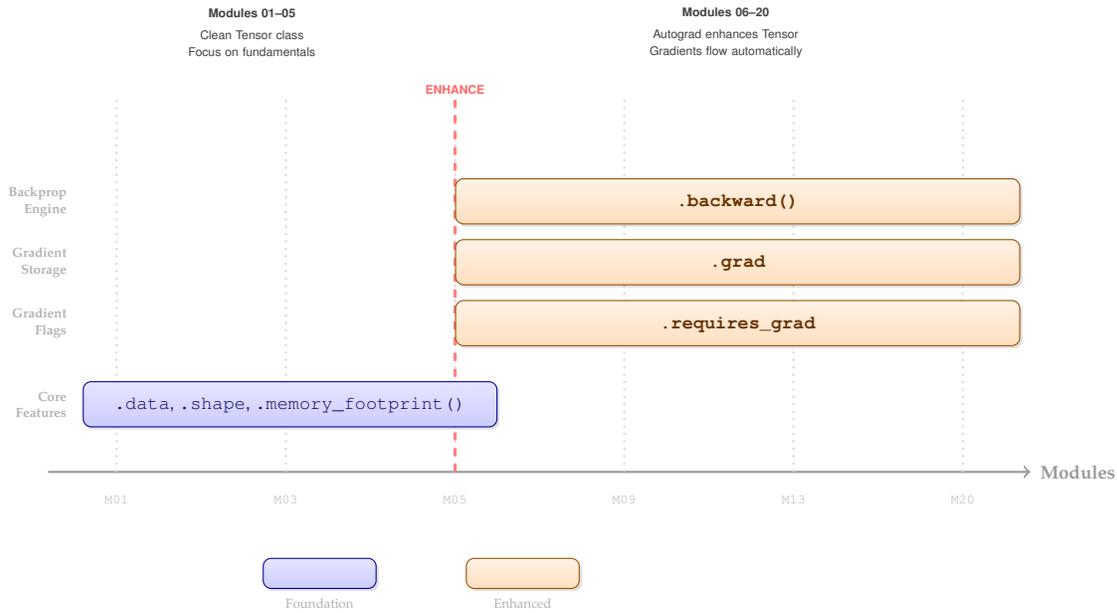
\begin{figure*}[t]
\centering
\begin{tikzpicture}[
    scale=0.9,
    every node/.style={font=\scriptsize\sffamily},
    foundation/.style={
        rectangle, 
        draw=blue!60!black, 
        top color=blue!10, 
        bottom color=blue!20, 
        text=blue!40!black, 
        minimum width=2.0cm, 
        minimum height=0.6cm, 
        anchor=east,
        rounded corners=3pt,
        drop shadow={opacity=0.25, shadow xshift=1pt, shadow yshift=-1pt}
    },
    enhanced/.style={
        rectangle, 
        draw=orange!60!black, 
        top color=orange!10, 
        bottom color=orange!25, 
        text=orange!40!black, 
        font=\scriptsize\bfseries\sffamily, 
        minimum width=2.0cm, 
        minimum height=0.6cm, 
        anchor=west,
        rounded corners=3pt,
        drop shadow={opacity=0.25, shadow xshift=1pt, shadow yshift=-1pt}
    },
    timeline/.style={thick, gray!80, line cap=round}
]

\draw[timeline, ->] (0,0) -- (14.5,0) node[right, font=\scriptsize\bfseries, gray!80] {Modules};

\foreach \x/\label in {1/01, 3.5/03, 6/05, 8.5/09, 11/13, 13.5/20} {
    \draw[gray!30, dotted, thick] (\x, 0) -- (\x, 5.5);
    \node[below, font=\tiny\sffamily, gray!60] at (\x, -0.2) {\texttt{M\label}};
}

\draw[red!50, very thick, dashed] (6, 0) -- (6, 5.5);
\node[above, font=\tiny\bfseries\sffamily, red!60, fill=white, inner sep=2pt, rounded corners=2pt] at (6, 5.5) {ENHANCE};

\node[foundation, minimum width=5.5cm, anchor=west] at (0.5, 1.0) {\texttt{.data}, \texttt{.shape}, \texttt{.memory\_footprint()}};
\node[left, font=\tiny\bfseries, gray!60, align=right] at (0.4, 1.0) {Core\\Features};

\node[enhanced, minimum width=7.5cm] at (6, 2.2) {\texttt{.requires\_grad}};
\node[left, font=\tiny\bfseries, gray!60, align=right] at (0.4, 2.2) {Gradient\\Flags};

\node[enhanced, minimum width=7.5cm] at (6, 3.1) {\texttt{.grad}};
\node[left, font=\tiny\bfseries, gray!60, align=right] at (0.4, 3.1) {Gradient\\Storage};

\node[enhanced, minimum width=7.5cm] at (6, 4.0) {\texttt{.backward()}};
\node[left, font=\tiny\bfseries, gray!60, align=right] at (0.4, 4.0) {Backprop\\Engine};

\node[align=center, font=\tiny\sffamily, text width=4.5cm, gray!50!black] at (3, 6.5) {
    \textbf{Modules 01--05}\\[2pt]
    Clean Tensor class\\
    Focus on fundamentals
};

\node[align=center, font=\tiny\sffamily, text width=4.5cm, gray!50!black] at (10, 6.5) {
    \textbf{Modules 06--20}\\[2pt]
    Autograd enhances Tensor\\
    Gradients flow automatically
};

\begin{scope}[shift={(4, -1.5)}]
    \node[foundation, minimum width=1.5cm, minimum height=0.4cm, anchor=center, label={[font=\tiny, gray!60]below:Foundation}] at (0, 0) {};
    \node[enhanced, minimum width=1.5cm, minimum height=0.4cm, anchor=center, label={[font=\tiny, gray!60]below:Enhanced}] at (3, 0) {};
\end{scope}

\end{tikzpicture}
\caption{\textbf{Progressive Disclosure Timeline.} Runtime capability expansion manages cognitive load. Modules 01--05 use a clean Tensor class focused on fundamentals (blue): data storage, arithmetic, matrix operations, and memory profiling. In Module 06, \texttt{enable\_autograd()} enhances Tensor with gradient tracking (orange): \texttt{requires\_grad}, \texttt{.grad}, and \texttt{.backward()} are dynamically added via monkey-patching. Three learning benefits: (1) students master fundamentals without distraction from unused features; (2) Module 01--05 code continues working unchanged after enhancement (backward compatibility); (3) the enhancement moment in Module 06 becomes a concrete ``aha'' experience as familiar tensors gain new capabilities.}
\label{fig:progressive-timeline}
\end{figure*}

TinyTorch's Module 01 \texttt{Tensor} class focuses exclusively on core tensor operations: data storage, arithmetic, matrix multiplication, and shape manipulation (\Cref{lst:foundation-tensor}). No gradient-related attributes exist yet, so students learn tensor fundamentals without cognitive overhead from features they won't use for five more modules. In Module 06, the \texttt{enable\_autograd()} function dynamically enhances \texttt{Tensor} with gradient tracking capabilities through monkey-patching (\Cref{lst:activation}). \Cref{fig:progressive-timeline} visualizes this enhancement timeline across the curriculum.

\begin{lstlisting}[caption={\textbf{Foundation Tensor.} Module 01 Tensor focuses on core operations. No gradient infrastructure exists yet; students learn tensor fundamentals first.},label={lst:foundation-tensor},float=t]
# Module 01: Foundation Tensor
class Tensor:
  def __init__(self, data):
    self.data = np.array(data, dtype=np.float32)
    self.shape = self.data.shape
    self.size = self.data.size
    self.dtype = self.data.dtype
    # No gradient features - pure data container

  def memory_footprint(self):
    """Systems thinking from day one"""
    return self.data.nbytes

  def __mul__(self, other):
    return Tensor(self.data * other.data)
\end{lstlisting}

\begin{lstlisting}[caption={\textbf{Autograd Enhancement.} Module 06 monkey-patches Tensor to add gradient tracking. The original \texttt{\_\_init\_\_} is wrapped to accept \texttt{requires\_grad}, and operations are enhanced to build computation graphs.},label={lst:activation},float=t]
def enable_autograd():
  """Enhance Tensor with gradient tracking"""
  _original_init = Tensor.__init__

  def gradient_aware_init(self, data, requires_grad=False):
    _original_init(self, data)
    self.requires_grad = requires_grad
    self.grad = None

  Tensor.__init__ = gradient_aware_init
  # Also patch __add__, __mul__, etc. to track gradients
  print("Autograd enabled!")

# Module 06 auto-enables on import
enable_autograd()
x = Tensor([3.0], requires_grad=True)
y = x * x  # y = 9.0
y.backward()
print(x.grad)  # [6.0] - dy/dx = 2x
\end{lstlisting}

\subsection{Pedagogical Justification}

Progressive disclosure is grounded in cognitive load theory~\citep{sweller1988cognitive} and threshold concept pedagogy~\citep{meyer2003threshold}. By deferring gradient infrastructure until Module 06, students focus entirely on tensor fundamentals without the cognitive overhead of unused attributes. Autograd represents a threshold concept (transformative and troublesome) introduced only when students have mastered the prerequisites (forward pass, loss computation) needed to appreciate it. The enhancement moment creates a concrete learning experience: tensors students already understand suddenly gain powerful new capabilities.

\textbf{Implementation Choice: Monkey-Patching vs. Inheritance.} Alternative designs include inheritance (\texttt{TensorV1}/\texttt{TensorV2}) or composition. We chose monkey-patching because it maintains a single \texttt{Tensor} class throughout the curriculum, mirrors PyTorch 0.4's Variable-Tensor merger via runtime consolidation, and creates a memorable ``activation moment'' when capabilities appear. Students implement the \texttt{enable\_autograd()} function themselves, learning Python metaprogramming (runtime class modification) as a systems concept, the same pattern used in production frameworks for backward compatibility and feature flags. The software engineering trade-off (global state modification) is explicitly discussed in Module 06's reflection questions.

\subsection{Production Framework Alignment}

Progressive disclosure demonstrates how real ML frameworks evolve. Early PyTorch (pre-0.4) separated data (\texttt{torch.Tensor}) from gradients (\texttt{torch.autograd.Variable}). PyTorch 0.4 (April 2018) \citep{pytorch04release} consolidated functionality into \texttt{Tensor}, matching TinyTorch's pattern. Students are exposed to the modern unified interface from Module 01, positioned to understand why PyTorch made this design evolution.

Similarly, TensorFlow 2.0 integrated eager execution by default \citep{tensorflow20}, making gradients work immediately (similar to TinyTorch's activation pattern). Students who understand progressive disclosure can grasp why TensorFlow eliminated \texttt{tf.Session()}: immediate execution with automatic graph construction aligns with unified API design principles.

\section{Systems-First Integration}
\label{sec:systems}

Having established TinyTorch's systems-first architecture (\Cref{sec:curriculum}), this section details how systems awareness manifests through a three-phase progression: (1) \textbf{understanding memory} through explicit profiling, (2) \textbf{analyzing complexity} through transparent implementations, and (3) \textbf{optimizing systems} through measurement-driven iteration. This progression applies situated cognition \citep{lave1991situated} by mirroring professional ML engineering workflow: measure resource requirements, understand computational costs, then optimize bottlenecks.

\subsection{Phase 1: Understanding and Characterizing Memory Usage}

Where traditional frameworks abstract away memory concerns, TinyTorch makes memory footprint calculation explicit (\Cref{lst:tensor-memory}). Students calculate memory footprints, discovering that a single batch of 32 ImageNet images requires 19 MB, while the full dataset exceeds 670 GB. This memory-first pedagogy transforms student questions:

\begin{itemize}
\item Module 01: ``Why does batch size affect memory?'' (activations scale with batch size)
\item Module 06: ``Why does Adam use 2$\times$ optimizer state memory?'' (momentum and variance buffers)
\item Module 13: ``How much VRAM for GPT-3 training?'' (175B parameters $\times$ 4 bytes $\times$ 4 $\approx$ 2.6 TB: weights + gradients + momentum + variance)
\end{itemize}

Students learn to distinguish parameter memory (model weights) from optimizer state memory from activation memory (often 10--100$\times$ larger than parameters). This decomposition enables accurate capacity planning for training runs.

\subsection{Phase 2: Analyzing Complexity Through Transparent Implementations}

Module 09 introduces convolution with seven explicit nested loops (\Cref{lst:conv-explicit}), making $O(B \times C_{\text{out}} \times H_{\text{out}} \times W_{\text{out}} \times C_{\text{in}} \times K_h \times K_w)$ complexity visible and countable.

\begin{lstlisting}[caption={\textbf{Explicit Convolution.} Seven nested loops reveal $O(C_{out} \times H \times W \times C_{in} \times K^2)$ complexity.},label={lst:conv-explicit},float=t]
def conv2d_explicit(input, weight):
  """7 nested loops - see the complexity!
  input: (B, C_in, H, W)
  weight: (C_out, C_in, K_h, K_w)"""
  B, C_in, H, W = input.shape
  C_out, _, K_h, K_w = weight.shape
  H_out, W_out = H - K_h + 1, W - K_w + 1
  output = np.zeros((B, C_out, H_out, W_out))

  # Count: 1,2,3,4,5,6,7 loops
  for b in range(B):
    for c_out in range(C_out):
      for h in range(H_out):
        for w in range(W_out):
          for c_in in range(C_in):
            for kh in range(K_h):
              for kw in range(K_w):
                output[b,c_out,h,w] += \
                  input[b,c_in,h+kh,w+kw] * \
                  weight[c_out,c_in,kh,kw]
  return output
\end{lstlisting}

This explicit implementation illustrates TinyTorch's pedagogical philosophy of minimal NumPy reliance until concepts are established. While the curriculum builds on NumPy as foundational infrastructure (array storage, broadcasting, element-wise operations), optimized operations like matrix multiplication appear only after students understand computational complexity through explicit loops. Module 03 introduces linear layers with manual weight-input multiplication loops before Module 08 introduces NumPy's \texttt{@} operator; Module 09 teaches convolution through seven nested loops before Module 17 vectorizes with NumPy operations. This progression ensures students understand \emph{what} operations do (and their complexity) before learning \emph{how} to optimize them. Pure Python transparency enables this pedagogical sequencing: students can inspect every operation without navigating compiled C extensions or CUDA kernels.

Students calculate: CIFAR-10 batch (128, 3, 32, 32) through 32-filter 5$\times$5 convolution: $128 \times 32 \times 28 \times 28 \times 3 \times 5 \times 5 = 241$M multiply-accumulate operations. This concrete measurement motivates Module 17's vectorization (10--100$\times$ speedup) and explains why CNNs require hardware acceleration.

\textbf{Experiencing Performance Reality.} TinyTorch's pure Python implementations are deliberately 100--10,000$\times$ slower than PyTorch. This slowness is pedagogically valuable (productive failure \citep{kapur2008productive}): students experience performance problems before learning optimizations, making vectorization meaningful rather than abstract. When students measure their Conv2d taking 97 seconds per CIFAR batch versus PyTorch's 10 milliseconds, they understand \emph{why} production frameworks obsess over implementation details.

\subsection{Phase 3: Optimizing Systems Through Measurement-Driven Iteration}

The Optimization Tier (Modules 14--19) transforms systems-first pedagogy from \emph{analysis} (``How much memory does this use?'') into \emph{optimization} (``How do I reduce memory by 4$\times$?''). Where foundation modules taught calculating footprints and counting FLOPs, optimization modules teach systematic improvement through profiling-driven iteration.

This tier introduces three fundamental optimization concepts that complete the systems-first integration:

\textbf{1. Measurement-Driven Optimization.} Students learn the ``measure first, optimize second'' methodology through systematic profiling. Rather than guessing bottlenecks, they measure time, memory, and FLOPs to identify where optimization efforts yield maximum impact. This mirrors production ML engineering: profiling reveals that convolution consumes 80\%+ training time, directing optimization focus appropriately.

\textbf{2. Trade-off Reasoning.} Optimization involves balancing competing objectives: accuracy, speed, memory, model size. Students measure these trade-offs (quantization achieves 4$\times$ compression with 1--2\% accuracy cost; pruning removes 90\% of parameters with minimal accuracy impact; KV-caching achieves 10--100$\times$ speedup but increases memory). This reinforces that systems engineering requires navigating trade-offs, not absolutes.

\textbf{3. Implementation Matters.} Identical algorithms exhibit 100$\times$ performance differences based on implementation choices. Students experience this through vectorization: seven explicit loops (pedagogically transparent) versus NumPy matrix operations (production efficient). This teaches why production frameworks obsess over seemingly minor implementation details: performance differences compound across operations.

The Optimization Tier completes the systems-first integration arc: students who calculate memory in Module 01, count FLOPs in Module 09, and optimize deployment in Modules 14--19 are designed to develop reflexive systems thinking. When encountering new ML techniques, the curriculum aims to train them to automatically ask: ``How much memory? What's the computational complexity? What are the trade-offs?'' Whether this design successfully makes these questions automatic rather than afterthoughts requires empirical validation.

\section{Course Deployment}
\label{sec:deployment}

Translating curriculum design into effective classroom practice requires addressing integration models, infrastructure accessibility, and student support structures. This section presents deployment patterns designed for diverse institutional contexts.

\textbf{Textbook Integration.}
TinyTorch serves as the hands-on implementation companion to the \emph{Machine Learning Systems} textbook~\citep{mlsysbook2025} (\texttt{mlsysbook.ai}), creating synergy between theoretical foundations and systems engineering practice. While the textbook covers the full ML lifecycle—data engineering, training architectures, deployment monitoring, robust operations, and sustainable AI—TinyTorch provides the complementary experience of building core infrastructure from first principles. This integration enables a complete educational pathway: students study production ML systems architecture in the textbook (Chapter 4: distributed training patterns, Chapter 7: quantization strategies), then implement those same abstractions in TinyTorch (Module 06: autograd for backpropagation, Module 15: INT8 quantization). The two resources address different aspects of the same educational gap: understanding both \emph{how production systems work} (textbook's systems architecture perspective) and \emph{how to build them yourself} (TinyTorch's implementation depth).

\subsection{Integration Models}
\label{subsec:integration}

TinyTorch supports three deployment models for different institutional contexts, ranging from standalone courses to supplementary tracks in existing curricula.

\textbf{Model 1: Self-Paced Learning (Primary Use Case)} serves individual learners, professionals, and researchers wanting framework internals understanding. Students work through modules at their own pace, selecting depth based on goals: complete all 20 modules for comprehensive systems knowledge, focus on Foundation (01--08) for autograd understanding, or target specific topics (Module 12 for attention mechanisms, Module 15 for quantization). The curriculum provides immediate feedback through local NBGrader validation, historical milestones for correctness proof, and progressive complexity enabling both intensive study (weeks) and distributed learning (months). This model requires zero infrastructure beyond Python and 4GB RAM, making it accessible worldwide.

\textbf{Model 2: Institutional Integration} enables universities to incorporate TinyTorch into existing ML courses. Options include: standalone 4-credit course (all 20 modules, complete systems coverage), half-semester module (Modules 01--09, foundation + CNN architectures), or optional honors track (selected modules for extra credit). Institutional deployment provides NBGrader autograding infrastructure, connection maps showing prerequisite dependencies, and milestone validation scripts. Lecture materials remain future work; current release supports lab-based or flipped-classroom formats where students implement concepts from textbook readings.

\textbf{Model 3: Team Onboarding} addresses industry use cases where ML teams want members to understand PyTorch internals. Companies can use TinyTorch for: new hire bootcamps (2--3 week intensive), internal training programs (distributed over quarters), or debugging workshops (focused modules like 06 Autograd, 12 Attention). The framework's PyTorch-inspired package structure and systems-first approach prepare engineers for understanding production frameworks and optimization workflows.

\textbf{Available Resources}: Current release provides module notebooks, NBGrader test suites, milestone validation scripts, and connection maps. Lecture slides for institutional courses remain future work (\Cref{sec:future-work}), though self-paced learning requires no additional materials beyond the modules themselves.

\subsection{Tier-Based Curriculum Configurations}
\label{subsec:tier-configs}

TinyTorch's three-tier architecture (Foundation, Architecture, Optimization) enables flexible deployment matching diverse course objectives and time constraints. Instructors can deploy complete tiers or selectively focus on specific learning goals:

\textbf{Configuration 1: Foundation Only (Modules 01--08).} Students build core framework internals from scratch: tensors, activations, layers, losses, dataloader, autograd, optimizers, and training loops. This configuration suits introductory ML systems courses, undergraduate capstone projects, or bootcamp modules focusing on framework fundamentals. Students complete Milestones 1--3 (Perceptron, XOR, MLP Revival) demonstrating functional autograd and training infrastructure. Upon completion, students understand \texttt{loss.backward()} mechanics, can debug gradient flow, and profile memory usage. Ideal for courses prioritizing systems fundamentals over architectural breadth.

\textbf{Configuration 2: Foundation + Architecture (Modules 01--13).} Extends Configuration 1 with modern deep learning architectures: datasets/dataloaders, convolution, pooling, embeddings, attention, and transformers. This configuration enables comprehensive ML systems courses or graduate-level deep learning seminars. Students complete Milestones 4--5 (CNN Revolution, Transformer Era) demonstrating working vision and language models. Upon completion, students implement production architectures from scratch, understand memory scaling ($O(N^2)$ attention), and recognize architectural tradeoffs (109$\times$ parameter efficiency from Conv2d weight sharing). Suitable for semester-long courses covering both internals and modern ML.

\textbf{Configuration 3: Optimization Focus (Modules 14--19 only).} Students import pre-built \texttt{tinytorch.core} modules (layers, optimizers, spatial operations) from Configurations 1--2, implementing only production optimization techniques: profiling, quantization, compression, acceleration, memoization, and benchmarking. This configuration targets production ML courses, TinyML workshops, or edge deployment seminars where students already understand framework basics but need systems optimization depth. Students complete Milestone 6 (MLPerf Benchmarks capstone) demonstrating reproducible benchmarking and optimization workflows. Upon completion, students optimize existing models for deployment constraints. Addresses key pedagogical limitation: students interested in quantization shouldn't need to re-implement autograd first.

These configurations support "build what you're learning, import what you need" pedagogy. Configuration 3 students focus on optimization while treating Foundation/Architecture as trusted dependencies, mirroring professional practice where engineers specialize rather than rebuilding entire stacks. The three-tier structure also enables multi-semester deployments aligned with academic terms, and hybrid integration where TinyTorch modules augment PyTorch-first courses by revealing framework internals (e.g., implementing Module 06 autograd to understand \texttt{loss.backward()}, or Module 09 convolution to demystify \texttt{torch.nn.Conv2d}).

\subsection{ML Systems Competency Coverage}
\label{subsec:coverage}

The ML Systems Competency Matrix (\Cref{tab:competency-matrix}) established 40 cells as the design framework. \Cref{tab:coverage} maps TinyTorch's 20 modules to this framework, demonstrating systematic coverage across knowledge areas and capability levels.

\begin{table*}[t]
\centering
\caption{\textbf{ML Systems Competency Coverage.} Each cell shows which modules address that competency. \fullmark~= full coverage through implementation and assessment. \halfmark~= partial coverage through conceptual instruction. \emptymark~= intentional gap where production infrastructure exceeds accessibility-first scope. Of 40 cells, 36 receive full coverage, 3 receive partial coverage, and 1 represents an intentional gap.}
\label{tab:coverage}
\small
\renewcommand{\arraystretch}{1.4}
\setlength{\tabcolsep}{6pt}
\begin{tabularx}{0.98\textwidth}{@{}>{\raggedright\arraybackslash}p{2.4cm}>{\raggedright\arraybackslash}X>{\raggedright\arraybackslash}X>{\raggedright\arraybackslash}X>{\raggedright\arraybackslash}X>{\raggedright\arraybackslash}X@{}}
\toprule
\textbf{Knowledge Area} & \textbf{Knows} & \textbf{Measures} & \textbf{Implements} & \textbf{Debugs} & \textbf{Optimizes} \\
\midrule
Tensors \& Memory & \fullmark~M01 & \fullmark~M01, M14 & \fullmark~M01 & \fullmark~M01 & \halfmark~M18 \\
\addlinespace[2pt]
Operations \& Compute & \fullmark~M01, M02 & \fullmark~M14 & \fullmark~M01, M02, M09, M12 & \fullmark~M02, M09 & \fullmark~M18 \\
\addlinespace[2pt]
Graphs \& Differentiation & \fullmark~M06 & \fullmark~M06, M14 & \fullmark~M06 & \fullmark~M06 & \halfmark~conceptual \\
\addlinespace[2pt]
Optimization \& Training & \fullmark~M07, M08 & \fullmark~M07, M14 & \fullmark~M07, M08 & \fullmark~M08 & \fullmark~M08 \\
\addlinespace[2pt]
Architectures \& Models & \fullmark~M03, M09--M13 & \fullmark~M14 & \fullmark~M03, M09--M13 & \fullmark~milestones & \fullmark~M16 \\
\addlinespace[2pt]
Pipelines \& Data & \fullmark~M05 & \fullmark~M05, M14 & \fullmark~M05 & \fullmark~M05 & \emptymark~CPU-only \\
\addlinespace[2pt]
Efficiency \& Compression & \fullmark~M15, M16 & \fullmark~M15, M16, M19 & \fullmark~M15, M16 & \fullmark~M15, M16 & \fullmark~M15, M16 \\
\addlinespace[2pt]
Deployment \& Serving & \fullmark~M17 & \fullmark~M19 & \fullmark~M17 & \halfmark~limited & \fullmark~M17 \\
\bottomrule
\end{tabularx}
\end{table*}

\textbf{Coverage Analysis.} TinyTorch provides full coverage of 36 cells (90\%), partial coverage of 3 cells (7.5\%), and an intentional gap in 1 cell (2.5\%). Full coverage means students implement working code, measure relevant metrics, and receive automated assessment. Partial coverage indicates conceptual instruction without complete implementation.

\textbf{Intentional Gap.} One cell remains uncovered by design: \textbf{Pipelines-Optimizes} (parallel data loading, GPU prefetch) requires multi-threaded I/O and GPU memory management beyond CPU-only scope. Students understand concepts through M05 but don't implement production-grade parallel loaders.

\textbf{Partial Coverage.} Three cells receive conceptual treatment: \textbf{Tensors-Optimizes} (memory pooling taught conceptually, students implement in-place operations but not allocator design), \textbf{Graphs-Optimizes} (activation checkpointing explained as memory-compute tradeoff but not implemented), and \textbf{Deployment-Debugs} (timeout debugging discussed without production environment practice). These represent deliberate scope boundaries. Students completing TinyTorch are prepared to address these areas through subsequent coursework or industry experience, building on strong foundations across the remaining 39 cells.

\subsection{Infrastructure and Accessibility}
\label{subsec:infrastructure}

ML systems education faces an accessibility challenge: production ML courses typically require expensive GPU hardware (\$500+ gaming laptops or cloud credits), 16GB+ RAM, CUDA-compatible environments, and Linux/WSL systems. These requirements create barriers for community college students, international learners in regions with limited cloud access, K-12 educators exploring ML internals, and institutions with modest computing budgets. Widening access to ML systems education requires reducing infrastructure barriers while maintaining pedagogical effectiveness~\citep{banbury2021widening}.

TinyTorch addresses this through CPU-only, pure Python implementation. The curriculum requires only dual-core 2GHz+ CPUs (no GPU needed), 4GB RAM (sufficient for CIFAR-10 training with batch size 32), 2GB storage (modules plus datasets), and any operating system supporting Python 3.8+ (Windows, macOS, or Linux). This enables deployment on Chromebooks via Google Colab, five-year-old budget laptops, and institutional computer labs. Text-based ASCII connection maps enhance accessibility for visually impaired students using screen readers, while offline-first datasets (\Cref{sec:curriculum}) eliminate network dependencies during training.

\subsubsection{Jupyter Environment Options}

TinyTorch supports three deployment environments: \textbf{JupyterHub} (institutional server, 8-core/32GB supports 50 students), \textbf{Google Colab} (zero installation, best for MOOCs), and \textbf{local installation} (\texttt{pip install tinytorch}, best for self-paced learning).

\subsubsection{NBGrader Autograding Workflow}

\textbf{Student Submission Process}: (1) Student works in Jupyter notebook (local or cloud), (2) runs \texttt{nbgrader validate module\_01.ipynb} for local correctness checking, (3) submits via LMS (Canvas/Blackboard) or Git (GitHub Classroom), (4) instructor runs \texttt{nbgrader autograde} on submitted notebooks, (5) grades and feedback posted to LMS.

\textbf{NBGrader Module Structure Example}: Each module uses NBGrader markdown cells to define assessment points and structure. For example, Module 01's memory profiling exercise:

\begin{lstlisting}[caption={\textbf{NBGrader Structure.} Cell metadata defines point allocation and solution delimiters.},label={lst:nbgrader-example},float=t]
# Cell metadata defines grading parameters:
# nbgrader = {
#   "grade": true,
#   "grade_id": "tensor_memory",
#   "points": 2,
#   "locked": false,
#   "solution": true
# }

def memory_footprint(self):
  """Calculate tensor memory in bytes"""
  ### BEGIN SOLUTION
  return self.data.nbytes
  ### END SOLUTION
\end{lstlisting}

This scaffolding~\citep{vygotsky1978mind} makes educational objectives explicit while enabling automated grading. The \texttt{name} field identifies the exercise, \texttt{points} assigns weight, and the description provides context before students see code cells.

\textbf{Handling Autograder Edge Cases}: Pure Python convolution (Module 09) may exceed default 30-second timeout on slower hardware; we set 5-minute timeouts and provide vectorized reference solutions for comparison. Critical modules (06 Autograd, 09 CNNs) include manual review of 20\% of submissions to catch conceptual errors missed by unit tests. All modules include \texttt{assert numpy.\_\_version\_\_ >= '1.20'} dependency validation.

\textbf{Projected Scalability}: Small courses (30 students) can grade in approximately 10 minutes per module on instructor laptop, medium courses (100 students) require approximately 30 minutes on dedicated grading server, while MOOCs (1000+ students) can achieve 2-hour turnaround via parallelized cloud autograding. These projections assume average grading time of 45 seconds per module submission on 4-core systems. Full-scale deployment validation planned for Fall 2025 (\Cref{sec:discussion}).

\subsection{Automated Assessment Infrastructure}

TinyTorch integrates NBGrader~\citep{blank2019nbgrader} for scalable automated assessment~\citep{thompson2008bloom}. Each module contains \textbf{solution cells} (scaffolded implementations with grade metadata), \textbf{test cells} (locked autograded tests preventing modification), and \textbf{point allocations} reflecting pedagogical priorities (Module 06 Autograd: 100 points; Module 01 Tensor: 60 points). Students validate correctness locally before submission, enabling immediate feedback.

This infrastructure enables deployment in MOOCs and large classrooms where manual grading proves infeasible. Instructors configure NBGrader to collect submissions, execute tests in sandboxed environments, and generate grade reports automatically.

\subsection{Package Organization}
\label{subsec:package}

Unlike tutorial-style notebooks creating isolated code, TinyTorch modules export to a package structure inspired by PyTorch's API organization. Each completed module becomes immediately usable: students build a working framework progressively, not isolated exercises. Module 01 exports to \texttt{tinytorch.core.tensor}, Module 09 to \texttt{tinytorch.core.spatial}, enabling import patterns familiar to PyTorch users that grow with each module completed.

As students complete modules, their framework accumulates capabilities. After Module 03, students can import and use layers; after Module 06, autograd enables training; after Module 09, CNNs become available. This progressive accumulation creates tangible evidence of progress: students see their framework grow from basic tensors to a complete ML system. \Cref{lst:progressive-imports} illustrates how imports expand as modules are completed:

\begin{lstlisting}[caption={\textbf{Progressive Imports.} Framework capabilities grow module-by-module as students complete implementations.},label={lst:progressive-imports},float=t]
# After Module 01: Basic tensors
from tinytorch import Tensor

# After Module 09: CNNs available
from tinytorch import Conv2d, MaxPool2d, Linear
# Autograd active - gradients flow!

# After Module 13: Complete framework
from tinytorch import GPT, Embedding, MultiHeadAttention
\end{lstlisting}

This design bridges educational and professional contexts. Students aren't ``solving exercises.'' They're building a framework they could ship. The package structure reinforces systems thinking: understanding how \texttt{tinytorch.Conv2d} relates to \texttt{tinytorch.Tensor} requires grasping module organization, not just individual algorithms. More importantly, students experience the satisfaction of watching their framework grow from a single \texttt{Tensor} class to a complete system capable of training transformers: each module completion adds new capabilities they can immediately use.

TinyTorch implements a literate programming workflow where source files serve dual purposes: executable Python code and educational documentation. Export happens via nbdev~\citep{howard2020fastai} directives (\texttt{\#| export}) embedded in module source files, enabling automatic package generation via \texttt{nbdev\_export}. TinyTorch modules are developed as Python source files using Jupytext percent format (\texttt{src/*/*.py}), with Jupyter notebooks generated for student distribution. The build system maintains single source of truth: developers edit \texttt{src/*/*.py} literate programming files containing both code and documentation, nbdev exports marked functions to \texttt{tinytorch/*} package structure (gitignored as generated artifact), and Jupytext converts source files to student-facing \texttt{.ipynb} notebooks. All 20 modules use \texttt{\#| export} directives for automatic package generation, enabling students to import from \texttt{tinytorch.core} and \texttt{tinytorch.perf} as they complete modules. This resolves the tension between version-controllable development (Python files enable proper diffs, merges, and code review) and notebook-based learning (students work in familiar Jupyter environments). Educators building similar curricula can adopt this pattern: maintain source-of-truth in version-controlled Python files while delivering interactive notebooks to students.

\subsection{Open Source Infrastructure}
\label{subsec:opensource}

TinyTorch is released as open source to enable community adoption and evolution.\footnote{Code released under MIT License, curriculum materials under Creative Commons Attribution-ShareAlike 4.0 (CC-BY-SA). Repository: \url{https://github.com/harvard-edge/cs249r_book/tree/main/tinytorch}} The repository includes instructor resources: \texttt{CONTRIBUTING.md} (guidelines for bug reports and curriculum improvements) and \texttt{INSTRUCTOR.md} (setup guide, grading rubrics, common student errors, and TA preparation strategies).

\textbf{Maintenance Commitment}: The author commits to bug fixes and dependency updates through 2027, community pull request review within 2 weeks, and annual releases incorporating educator feedback. Community governance transition (2026--2027) will establish an educator advisory board and document succession planning to ensure long-term sustainability beyond single-author maintenance.

\textbf{Customization Support}: TinyTorch's modular design enables institutional adaptation: replacing datasets with domain-specific data (medical images, time series), adding modules (diffusion models, graph neural networks), adjusting difficulty through scaffolding modifications, or changing assessment approaches. Forks should maintain attribution (CC-BY-SA requirement) and ideally contribute improvements upstream.

\subsection{Teaching Assistant Support}
\label{subsec:ta-support}

Effective deployment requires structured TA support beyond instructor guidance.

\textbf{TA Preparation}: TAs should develop deep familiarity with critical modules where students commonly struggle—Modules 06 (Autograd), 09 (CNNs), and 13 (Transformers)—by completing these modules themselves and intentionally introducing bugs to understand common error patterns. The \texttt{INSTRUCTOR.md} file documents frequent student errors (gradient shape mismatches, disconnected computational graphs, broadcasting failures) and debugging strategies.

\textbf{Office Hour Demand Patterns}: Student help requests are expected to cluster around conceptually challenging modules, with autograd (Module 06) likely generating higher office hour demand than foundation modules. Instructors should anticipate demand spikes by scheduling additional TA capacity during critical modules, providing pre-recorded debugging walkthroughs, and establishing async support channels (discussion forums with guaranteed response times).

\textbf{Grading Infrastructure}: While NBGrader automates 70-80\% of assessment, critical modules benefit from manual review of implementation quality and conceptual understanding. TAs should focus manual grading on: (1) code clarity and design choices, (2) edge case handling, (3) computational complexity analysis, and (4) memory profiling insights. Sample solutions and grading rubrics in \texttt{INSTRUCTOR.md} calibrate evaluation standards.

\textbf{Boundaries and Scaffolding}: TAs should guide students toward solutions through structured debugging questions rather than providing direct answers. When students reach unproductive frustration, TAs can suggest optional scaffolding modules (numerical gradient checking before autograd implementation, scalar autograd before tensor autograd) to build confidence through intermediate steps.

\subsection{Student Learning Support}
\label{subsec:student-support}

TinyTorch embraces productive failure \citep{kapur2008productive}, learning through struggle before instruction, while providing guardrails against unproductive frustration.

\textbf{Recognizing Productive vs Unproductive Struggle}: Productive struggle involves trying different approaches, making incremental progress (passing additional tests), and developing deeper understanding of error messages. Unproductive frustration manifests as repeated identical errors without new insights, random code changes hoping for success, or inability to articulate the problem. Students experiencing unproductive frustration should seek help rather than persisting solo.

\textbf{Structured Help-Seeking}: The repository provides debugging workflows: (1) self-debug using print statements and simple test cases, (2) consult common errors documentation for the module, (3) search discussion forums for similar issues, (4) post structured help requests with error messages and attempted solutions, (5) attend office hours with specific questions. This progression encourages independence while ensuring timely intervention.

\textbf{Flexible Pacing and Optional Scaffolding}: Students learn at different rates depending on background, learning style, and external commitments. TinyTorch supports multiple pacing modes (intensive weeks, semester distributed coursework, self-paced professional development) without prescriptive timelines. Students struggling with conceptual jumps can access optional intermediate modules providing additional scaffolding. No penalty attaches to slower pacing or scaffolding use; depth of understanding matters more than completion speed.

\textbf{Diverse Student Contexts}: The curriculum acknowledges students balance learning with work, caregiving, or health challenges. Flexible pacing enables participation from community college students, working professionals, international learners, and non-traditional students who might be excluded by rigid timelines or high-end hardware requirements. Pure Python deployment on modest hardware (4GB RAM, dual-core CPU) and screen-reader-compatible ASCII diagrams further broaden accessibility.

\section{Discussion and Limitations}
\label{sec:discussion}

This section reflects on TinyTorch's design through three lenses: pedagogical scope as deliberate design decision, flexible curriculum configurations enabling diverse institutional deployment, and honest assessment of limitations requiring future work.

\subsection{Pedagogical Scope as Design Decision}
\label{subsec:scope}

TinyTorch's single node, CPU-only, framework-internals-focused scope represents deliberate pedagogical constraint, not technical limitation. This scoping embodies three design principles:

\textbf{Accessibility over performance}: Pure Python eliminates GPU dependency, prioritizing equitable access over execution speed (pedagogical transparency detailed in \Cref{sec:systems}). GPU access remains inequitably distributed: cloud credits favor well-funded institutions, personal GPUs favor affluent students. The 100--10,000$\times$ slowdown versus PyTorch is acceptable when the pedagogical goal is understanding internals, not training production models.

\textbf{Incremental complexity management}: GPU programming introduces memory hierarchy (registers, shared memory, global memory), kernel launch semantics, race conditions, and hardware-specific tuning. Teaching GPU programming simultaneously with autograd would violate cognitive load constraints. TinyTorch enables a ``framework internals now, hardware optimization later'' learning pathway. Students completing TinyTorch should pursue GPU acceleration through PyTorch tutorials, NVIDIA Deep Learning Institute courses, or advanced ML systems courses, building on internals understanding to comprehend optimization techniques.

Similarly, distributed training (data parallelism, model parallelism, gradient synchronization) and production deployment (model serving, compilation, MLOps) introduce substantial additional complexity orthogonal to framework understanding. These topics remain important but beyond current pedagogical scope. Future extensions could address distributed systems through simulation-based pedagogy (\Cref{sec:future-work}), maintaining accessibility while teaching concepts.

\subsection{Limitations}

TinyTorch's current implementation contains gaps requiring future work.

\textbf{Experimentation constraints.} The performance trade-off discussed above (\Cref{subsec:scope}) limits practical experimentation. Students complete milestones (65--75\% CIFAR-10 accuracy, transformer text generation) but cannot iterate rapidly on architecture search or hyperparameter tuning.

\textbf{Energy consumption measurement.} While TinyTorch covers optimization techniques with significant energy implications (quantization achieving 4$\times$ compression, pruning enabling 10$\times$ model shrinkage), the curriculum does not explicitly measure or quantify energy consumption. Students understand that quantization reduces model size and pruning decreases computation, but may not connect these optimizations to concrete energy savings (joules/inference, watt-hours/training epoch). Future iterations could integrate energy profiling libraries to make sustainability an explicit learning objective alongside memory and latency optimization, particularly relevant for edge deployment.

\textbf{Language and accessibility.} Materials exist exclusively in English. Modular structure facilitates translation; community contributions welcome. Code examples omit type annotations (e.g., \texttt{def forward(self, x: Tensor) -> Tensor:}) to reduce visual complexity for students learning ML concepts simultaneously. While this prioritizes pedagogical clarity, it means students don't practice type-driven development increasingly standard in production ML codebases. Future iterations could introduce type hints progressively: omitting them in early Modules (01--05), then adding them in optimization Modules (14--18) where interface contracts become critical.

\textbf{Forward dependency prevention.} A recurring curriculum maintenance challenge is forward dependency creep: advanced concepts leaking into foundational modules through helper functions, error messages, or test cases that assume knowledge students haven't yet acquired. For example, an error message in Module 03 (Layers) that mentions ``computational graph'' assumes Module 06 (Autograd) knowledge. Maintaining pedagogical ordering requires vigilance during curriculum updates; future work could automate this validation through CI/CD checks that flag cross-module dependencies violating prerequisite ordering.

\section{Future Directions}
\label{sec:future-work}

TinyTorch's current implementation establishes a foundation for three extension directions: empirical validation to test pedagogical hypotheses, curriculum evolution to expand systems coverage beyond CPU-only scope, and community adoption to measure educational impact through deployment at scale.

\subsection{Empirical Validation}

While TinyTorch's design is grounded in established learning theory (cognitive load~\citep{sweller1988cognitive}, progressive disclosure, cognitive apprenticeship~\citep{collins1989cognitive}), its pedagogical effectiveness requires empirical validation through controlled classroom studies. Planned validation includes: (1) pilot deployments to measure cognitive load and identify curriculum refinements, (2) comparative studies testing whether build-from-scratch pedagogy improves conceptual understanding and debugging transfer versus application-first approaches, and (3) longitudinal tracking to assess retention and career outcomes.

\subsection{Curriculum Evolution}

TinyTorch deliberately focuses on single node ML systems fundamentals: understanding what happens inside \texttt{loss.backward()} on one machine, how memory flows through computational graphs, and why optimizer state consumes resources. This single node focus provides the foundation for understanding distributed systems, where the same operations must coordinate across multiple nodes with communication overhead, synchronization barriers, and failure modes that compound single node complexity. Future curriculum extensions would maintain TinyTorch's core principle (understanding through transparent implementation) while expanding to multi node systems coverage through simulation based pedagogy.

\textbf{Performance Analysis Through Analytical Models.} Future extensions could enable students to compare TinyTorch CPU implementations against PyTorch GPU equivalents through roofline modeling~\citep{williams2009roofline}. Rather than writing CUDA code, students would profile existing implementations to understand memory hierarchy differences, parallelism benefits, and compute versus memory bottlenecks. The roofline approach maintains TinyTorch's accessibility (no GPU hardware required) while preparing students for GPU programming by teaching first-principles performance analysis.

\textbf{Hardware Simulation Integration.} TinyTorch's current profiling infrastructure (memory tracking via tracemalloc, FLOP counting, and performance benchmarking) provides algorithmic-level analysis. Integrating architecture simulators (e.g., scale-sim~\citep{samajdar2018scale}, timeloop~\citep{parashar2019timeloop}, astra-sim~\citep{kannan2022astrasim}) would connect high-level ML operations with cycle-accurate hardware models. This layered approach mirrors real ML systems engineering: students first understand algorithmic complexity and memory patterns in TinyTorch, then trace those operations down to microarchitectural performance in simulators. Such integration would complete the educational arc from algorithmic implementation $\rightarrow$ systems profiling $\rightarrow$ hardware realization, maintaining TinyTorch's accessibility while preparing students for hardware-aware optimization.

\textbf{Energy and Power Profiling.} Edge deployment and sustainable ML~\citep{strubell2019energy,patterson2021carbon} require understanding energy consumption. Integrating power profiling tools would enable students to measure energy costs (joules per inference, watt-hours per training epoch) alongside latency and memory. This connects existing optimization techniques (quantization, pruning) taught in Modules 15--18 to concrete sustainability metrics, particularly relevant for edge AI~\citep{banbury2021benchmarking} where battery life constrains deployment.

\textbf{Distributed Training Through Simulation.} Understanding distributed training communication patterns requires simulation-based pedagogy rather than multi-GPU clusters. Leveraging architecture simulators, students could explore multi-node concepts: gradient synchronization overhead, scalability analysis across worker counts, network topology impact on communication patterns, and pipeline parallelism trade-offs. Students who master single-node fundamentals in TinyTorch would then explore how those same operations (forward pass, backward pass, optimizer step) change when distributed across nodes with communication latency and bandwidth constraints.

\textbf{Architecture Extensions.} Potential architecture additions (graph neural networks, diffusion models, reinforcement learning) must justify inclusion through systems pedagogy rather than completeness. The question is not ``Can TinyTorch implement this?'' but rather ``Does implementing this teach fundamental systems concepts unavailable through existing modules?'' Graph convolutions might teach sparse tensor operations; diffusion models might illuminate iterative refinement trade-offs. The curriculum remains intentionally incomplete as a production framework: completeness lies in foundational systems thinking applicable across all ML architectures.

\subsection{Community and Sustainability}

As part of the ML Systems Book ecosystem (\texttt{mlsysbook.ai}), TinyTorch benefits from broader educational infrastructure while the open-source model (MIT license) enables collaborative refinement across institutions: instructor discussion forums for pedagogical exchange, shared teaching resources, and empirical validation of learning outcomes.

As described earlier, Module 20 (Capstone) culminates the curriculum with competitive systems engineering challenges. Inspired by MLPerf benchmarking~\citep{mattson2020mlperf,reddi2020mlperf}, students optimize their implementations across accuracy, speed, compression, and efficiency dimensions, comparing results globally through standardized benchmarking infrastructure. This competitive element reinforces systems thinking: optimization requires measurement-driven decisions (profiling bottlenecks), principled tradeoffs (accuracy versus compression), and reproducible methodology (standardized metrics collection). The focus remains pedagogical (understanding \emph{why} optimizations work) rather than achieving state-of-the-art performance, but the competitive framing increases engagement and mirrors real ML engineering workflows.

\section{Conclusion}
\label{sec:conclusion}

Machine learning education faces a fundamental choice: teach students to \emph{use} frameworks as black boxes, or teach them to \emph{understand} what happens inside \texttt{loss.backward()}. TinyTorch demonstrates that deep systems understanding is accessible without GPU clusters or distributed infrastructure. \emph{Building systems creates irreversible understanding}: once you implement autograd, you cannot unsee the computational graph; once you profile memory allocation, you cannot unknow the costs. This accessibility matters: students worldwide can develop framework internals knowledge on modest hardware, transforming production debugging from trial-and-error into systematic engineering.

Three pedagogical contributions enable this transformation. Progressive disclosure manages complexity through gradual feature activation: students work with unified Tensor implementations that gain capabilities across modules rather than replacing code mid-semester. Systems-first integration embeds memory profiling from Module 01, preventing ``algorithms without costs'' learning where students optimize accuracy while ignoring deployment constraints. Historical milestone validation proves correctness through recreating nearly 70 years of ML breakthroughs (1958--2025, from Perceptron through Transformers), making abstract implementations concrete through reproducing published results.

\textbf{For ML practitioners}: Building TinyTorch's 20 modules transforms how you debug production failures. When PyTorch training crashes with OOM errors, you understand memory allocation across parameters, optimizer states, and activation tensors. When gradient explosions occur, you recognize backpropagation numerical instability from implementing it yourself. When choosing between Adam and SGD under memory constraints, you know the 4$\times$ total memory multiplier from building both optimizers. This systems knowledge transfers directly to production framework usage: you become an engineer who understands \emph{why} frameworks behave as they do, not just \emph{what} they do.

\textbf{For CS education researchers}: TinyTorch provides replicable infrastructure for testing pedagogical hypotheses about ML systems education. Does progressive disclosure reduce cognitive load compared to teaching autograd as separate framework? Does systems-first integration improve production readiness versus algorithms-only instruction? Do historical milestones increase engagement and retention? The curriculum embodies design patterns amenable to controlled empirical investigation. Open-source release with detailed validation roadmap enables multi-institutional studies to establish evidence-based best practices for teaching framework internals.

\textbf{For educators and bootcamp instructors}: TinyTorch supports flexible integration: self-paced learning requiring zero infrastructure (students run locally on laptops), institutional courses with automated NBGrader assessment, or industry team onboarding for ML engineers transitioning from application development to systems work. The modular structure enables selective adoption: foundation tier only (Modules 01--08, teaching core concepts), architecture focus (adding CNNs and Transformers through Module 13), or complete systems coverage (all 20 modules including optimization and deployment). No GPU access required, no cloud credits needed, no infrastructure barriers.

The complete codebase, curriculum materials, and assessment infrastructure are openly available at \texttt{mlsysbook.ai/tinytorch} (or \texttt{tinytorch.ai}) under permissive open-source licensing. We invite the global ML education community to adopt TinyTorch in courses, contribute curriculum improvements, translate materials for international accessibility, fork for domain-specific variants (robotics, edge AI), and empirically evaluate whether implementation-based pedagogy achieves its promise.

Sutton's Bitter Lesson teaches that general methods leveraging computation ultimately triumph, but someone must build the systems that enable that computation to scale. TinyTorch prepares students to be those engineers: the AI engineers who bridge the gap between what ML research makes possible and what reliable production systems require. They are practitioners who understand not just \emph{what} ML systems do, but \emph{why} they work and \emph{how} to make them scale.

\section*{Acknowledgments}

TinyTorch emerged from CS249r: Tiny Machine Learning at Harvard University, where teaching ML systems on constrained devices revealed that the same principles (memory management, computational efficiency, optimization tradeoffs) apply at every scale. The ``Tiny'' in TinyTorch reflects these origins, but the framework teaches general ML systems fundamentals that transfer from microcontrollers to datacenters. We thank the students across multiple offerings whose questions, struggles, and feedback shaped the curriculum into its current form. We also thank Colby Banbury and Zishen Wan for their feedback on this work, and the global community around \texttt{mlsysbook.ai} whose engagement continues to refine these materials.

\subsection*{Use of Generative AI}

In accordance with ACM policy on authorship, we disclose that generative AI tools were used during the development of both the TinyTorch framework and this manuscript, including code review, debugging assistance, proofreading, and improving clarity of exposition. We view this as aligned with TinyTorch's mission: just as TinyTorch aims to democratize ML systems education by removing infrastructure barriers, AI tools democratize the creation of educational resources by augmenting individual contributors. A curriculum of this scope would traditionally require a team; AI assistance enabled a single author to develop, test, and document 20 interconnected modules. The author takes full responsibility for all content, and the core intellectual contributions (curriculum structure, pedagogical framework, and systems first methodology) reflect human design decisions informed by teaching experience.

\bibliography{references}

\end{document}